\documentclass{article}
\usepackage[T1]{fontenc}

\usepackage{microtype}
\usepackage{graphicx}
\usepackage{booktabs} 
\usepackage{hyperref}
\usepackage{url}

\usepackage[accepted]{icml2024}

\usepackage[normalem]{ulem}
\usepackage{amssymb}
\usepackage{amsthm}
\usepackage{mathtools}
\usepackage{bm}
\usepackage{bbm}
\usepackage{dsfont}
\usepackage{xcolor}
\usepackage{accents}
\usepackage{cancel}
\usepackage{multicol}
\usepackage[english]{babel}
\renewcommand{\algorithmiccomment}[1]{\hfill \(\triangleright\) #1}

\definecolor{mypurple}{HTML}{332288}
\definecolor{mygreen}{HTML}{117733}
\definecolor{myfuchsia}{HTML}{882255}

\theoremstyle{plain}
\newtheorem{theorem}{Theorem}[section]

\newtheorem{lemma}{Lemma}[section]

\newcommand{\E}{\mathbb{E}}
\newcommand{\Real}{\mathbb{R}}

\newcommand{\st}{\mathbf{s}}
\newcommand{\St}{\mathbf{S}}
\newcommand{\Scal}{\mathcal{S}}

\newcommand{\at}{\mathbf{a}}
\newcommand{\At}{\mathbf{A}}
\newcommand{\Acal}{\mathcal{A}}

\newcommand{\rt}{\mathrm{r}}
\newcommand{\Rt}{\mathrm{R}}
\newcommand{\Rcal}{\mathcal{R}}

\newcommand{\Gt}{\mathrm{G}}

\newcommand{\Ht}{\mathrm{H}}

\newcommand{\step}{\mathrm{t}}
\newcommand{\Step}{\mathrm{T}}
\newcommand{\Tcal}{\mathcal{T}}

\newcommand{\Dcal}{\mathcal{D}}

\newcommand{\kstep}{\mathrm{k}}

\newcommand{\paramp}{\bm{\theta}}
\newcommand{\Paramp}{\bm{\Theta}}

\newcommand{\paramv}{\mathbf{w}}
\newcommand{\Paramv}{\mathbf{W}}

\newcommand{\Prob}{\mathbb{P}}
\newcommand{\prob}{\mathbb{P}}

\usepackage[capitalize,noabbrev]{cleveref}
\usepackage{caption}
\usepackage{subcaption}

\crefname{assumption}{assumption}{assumptions}

\ifdefined\nohyperref\else\ifdefined\hypersetup
  \definecolor{mydarkblue}{rgb}{0,0.08,0.45}
  \hypersetup{ %
    pdftitle={},
    pdfsubject={},
    pdfkeywords={},
    pdfborder=0 0 0,
    pdfpagemode=UseNone,
    colorlinks=true,
    linkcolor=mydarkblue,
    citecolor=mydarkblue,
    filecolor=mydarkblue,
    urlcolor=mydarkblue,
    }
  \fi
\fi

\icmltitlerunning{ReLU to the Rescue: Improve Your On-Policy Actor-Critic with Positive Advantages}

\begin{document}

\twocolumn[
\icmltitle{ReLU to the Rescue: Improve Your On-Policy Actor-Critic with Positive Advantages}



\icmlsetsymbol{equal}{*}

\begin{icmlauthorlist}
\icmlauthor{Andrew Jesson}{oatml}
\icmlauthor{Chris Lu}{flair}
\icmlauthor{Gunshi Gupta}{oatml}
\icmlauthor{Nicolas Beltran-Velez}{columbia}
\icmlauthor{Angelos Filos}{oatml}
\icmlauthor{Jakob N. Foerster}{flair}
\icmlauthor{Yarin Gal}{oatml}
\end{icmlauthorlist}

\icmlaffiliation{oatml}{OATML, Department of Computer Science, University of Oxford, Oxford, UK}
\icmlaffiliation{flair}{FLAIR, Department of Computer Science, University of Oxford, Oxford, UK}
\icmlaffiliation{columbia}{Department of Computer Science, Columbia University, New York, NY}

\icmlcorrespondingauthor{Andrew Jesson}{andrew.jesson@cs.ox.ac.uk}

\icmlkeywords{Machine Learning, ICML}

\vskip 0.3in
]



\printAffiliationsAndNotice{}

\begin{abstract}
    This paper proposes a step toward approximate Bayesian inference in on-policy actor-critic deep reinforcement learning. It is implemented through three changes to the Asynchronous Advantage Actor-Critic (A3C) algorithm: (1) applying a ReLU function to advantage estimates, (2) spectral normalization of actor-critic weights, and (3) incorporating \emph{dropout as a Bayesian approximation}. We prove under standard assumptions that restricting policy updates to positive advantages optimizes for value by maximizing a lower bound on the value function plus an additive term. We show that the additive term is bounded proportional to the Lipschitz constant of the value function, which offers theoretical grounding for spectral normalization of critic weights. Finally, our application of dropout corresponds to approximate Bayesian inference over both the actor and critic parameters, which enables \textit{adaptive state-aware} exploration around the modes of the actor via Thompson sampling. We demonstrate significant improvements for median and interquartile mean metrics over A3C, PPO, SAC, and TD3 on the MuJoCo continuous control benchmark and improvement over PPO in the challenging ProcGen generalization benchmark.
\end{abstract}

\section{Introduction}
\label{sec:introduction}

Deep Reinforcement Learning (DRL) finds approximate solutions to complex sequential decision-making problems in domains such as robotics \citep{ibarz2021train}, autonomous driving \citep{kiran2021deep}, strategy games \citep{mnih2015human,silver2017mastering,arulkumaran2019alphastar}, and human-computer interaction \citep{ziegler2019fine}. 
DRL algorithms achieve state-of-the-art performance on many challenging benchmarks \citep{young19minatar,gymnax2022github,todorov2012mujoco,brockman2016openai}.
However, their use in real-world applications depends on their capacity to execute tasks while making policy updates in the face of finite observations of a changing world. On-policy algorithms, such as Proximal Policy Optimization (PPO) \citep{schulman2017proximal} or Asynchronous Advantage Actor-Critic (A3C) \citep{mnih2016asynchronous}, update differentiable policies based on recent interactions with the environment. This recency bias and the capacity to actively sample informative observations make on-policy approaches compelling candidates for applications in dynamic real-world environments.

Exploration is a component of active sampling. 
On-policy actor-critic methods typically incorporate exploration through entropy regularization or by learning a homogeneous variance parameter for continuous action spaces \citep{schulman2015trust, mnih2016asynchronous, schulman2017proximal}. 
While effective, these exploration methods are \emph{state-agnostic}, promoting exploration equally regardless of the novelty or familiarity of a given state. 
Alternatively, the paradigm of maximum entropy reinforcement learning \citep{ziebart2008maximum,haarnoja2018soft,levine2018reinforcement} promotes \emph{state-aware} exploration through the inclusion of an actor entropy term in the optimization objective.
However, this term is not \emph{adaptive} in that it promotes higher actor entropy (more exploration) regardless of the number of state visits, and thus may be agnostic to the knowledge already gained about the system.

This work incorporates adaptive state-aware exploration into the on-policy actor-critic framework to improve the performance and efficiency of on-policy actor-critic algorithms.
Approximate Bayesian inference over actor weights would satisfy this goal, but its implementation is not straightforward due to the policy-gradient objective for optimizing the actor, as we will show. 
Thus, we ask, ``What is the minimal step we can take toward approximate Bayesian inference?'' and answer with VSOP (standing for Variational [b]ayes, Spectral-normalized, On-Policy reinforcement learning). 
VSOP consists of three simple modifications to the A3C algorithm: (1) applying a ReLU function to advantage estimates, (2) spectral normalization of actor-critic weights, and (3) incorporating \emph{dropout as a Bayesian approximation} \citep{gal2016dropout}. 
Under standard assumptions, we prove that restricting policy updates to positive advantages maximizes value by maximizing a lower bound on the value function plus an additive term. 
We show that the additive term is bounded proportional to the Lipschitz constant of the value function, grounding spectral normalization use as a Lipschitz constant regularizer. 
Finally, we show that approximate Bayesian inference over the actor and critic parameters enables adaptive state-aware exploration via Thompson sampling.

Through our thorough empirical assessments on the Gymnasium and Brax MuJoCo benchmarks for continuous control \citep{brockman2016openai,brax2021github}, we show that VSOP can significantly outperform existing DRL algorithms such as A3C, PPO, SAC, and TD3 for median and interquartile mean (IQM) metrics \citep{agarwal2021deep}. 
We further show through ablation studies that the union of our proposed improvements contributes significantly to increased performance and efficiency.
Finally, we show that VSOP significantly outperforms PPO on the challenging ProcGen generalization benchmark, demonstrating improved performance when deployed under distribution shift.

\section{Background}
\label{sec:background}

\textbf{Notation.}
We consider a discounted, $\Step$-horizon Markov Decision Process (MDP) defined by the tuple $(\Scal, \Acal, \mathrm{P}, \rt, \gamma)$, where $\Scal$ is the state space, $\Acal$ is the action space, $\mathrm{P}$ is the state transition probability, $\rt$ is the immediate reward upon transitioning from state $\st$ to state $\st^{\prime}$ under action $\at$, and $\gamma \in [0, 1]$ is the discount factor. MDPs provide a framework for modeling sequential decision-making problems, where an agent interacts with an environment over discrete time steps to achieve a goal \citep{puterman2014markov}. Following the notation of \citet{sutton2018reinforcement}, we define states at time $\step \in \Step$ by the $d$-dimensional, real-valued, random variable, $\St_{\step}: \Omega \to \Scal \subseteq \Real^d$, with observable instances $\st_{\step} = \St_{\step}(\omega_{\step}): \forall  \omega_{\step} \in \Omega$. We define actions by the $m$-dimensional random variable $\At_{\step}: \Omega \to \Acal$, with observable instances, $\at_{\step} = \At_{\step}(\omega_{\step}): \forall  \omega_{\step} \in \Omega$. Rewards are defined by the continuous-valued random variable, $\Rt_{\step}:\Omega \to \Rcal \subseteq \Real$, with observable instances, $\rt_{\step} = \Rt_{\step}(\omega_{\step}): \forall  \omega_{\step} \in \Omega$. Let the random variable, $\Gt_{\step} \coloneqq \sum_{\kstep = \step + 1}^{\Step} \gamma^{\kstep - 1 - \step} \Rt_{\kstep}$, denote the discounted return. We use the standard definitions for the conditional action distribution/density (policy), $\pi(\at \mid \st)$, the state value function under the policy, $v_{\pi}(\st) \coloneqq \E_{\pi} \left[ \Gt_{\step} \mid \St_{\step} = \st \right]$, and state-action value function under the policy, $q_{\pi}(\st, \at) \coloneqq \E_{\pi} \left[ \Gt_{\step} \mid \St_{\step} = \st,  \At_{\step} = \at\right]$.

\textbf{On-Policy Actor-Critic Reinforcement Learning.} 
On-policy, Actor-critic approaches to reinforcement learning are called \emph{policy-gradient} methods. They directly optimize a policy function, $\pi(\at \mid \st, \paramp)$, differentiable with respect to parameters, $\paramp$, to maximize the expected discounted return under the policy, $v_{\pi}(\st)$. On-policy approaches differ from off-policy approaches in that they only use recent samples from the current policy to achieve this objective. Actor-critic methods differ from other policy-gradient methods because they fit an approximate value function (critic), $v(\st, \paramv)$, to the data collected under the policy, in addition to optimizing the policy function (actor). The critic is typically used in actor optimization but not generally for decision-making.

Deep reinforcement learning implements the actor and critic using neural network architectures, where the function parameters correspond to network weights. We denote the parameters of the actor and critic networks as $\paramp$ and $\paramv$, respectively. The output likelihood of the actor makes distributional assumptions informed by characteristics of the action space, $\Acal$. A common choice for continuous action spaces is an independent multivariate normally distributed likelihood with homogeneous noise variance, $\pi(\at_{\step} \mid \st_{\step}, \paramp) \sim \mathcal{N}(\at \mid \bm{\mu}(\st, \paramp), \mathrm{I}\bm{\sigma}^2(\paramp))$, where $\bm{\sigma}^2(\paramp) = (\sigma^2_1, \dots, \sigma^2_m)$ is the vector of inferred action noise variances. For discrete action spaces, the likelihood is often a categorical distribution, $\pi(\at \mid \st, \paramp) \sim \mathrm{Categorical}(\at \mid \bm{\mu}(\st, \paramp))$. In both cases, the mean parameter of the likelihood, $\bm{\mu}(\st, \paramp)$, is the $m$-dimensional, vector-valued output of a neural network architecture with parameters, $\paramp$. Critic networks are commonly fit using a mean squared error objective, which implies a univariate normally distributed output likelihood with unit variance, $\mathrm{G} \mid \st, \paramv \sim \mathcal{N}(\mathrm{g} \mid v(\st, \paramv), 1)$, where the mean parameter is the approximate value function, $v(\st, \paramv)$, and is given by the scalar-valued output of any neural network architecture with parameters, $\paramv$.

The baseline on-policy, actor-critic policy gradient algorithm performs gradient ascent with respect to the ``performance'' function, $J(\paramp) \coloneqq v_{\pi}(\st_0, \paramp)$, where $v_{\pi}(\st_0, \paramp)$ is the value function with respect to the parameters $\paramp$ \citep{williams1992simple}. By the policy gradient theorem \citep{sutton1999policy}, we have: $\nabla_{\paramp} J(\paramp) = \nabla_{\paramp} v_{\pi}(\st_{0}) \propto \int_{\Scal} \rho(\st) \int_{\Acal} q_{\pi}(\st, \at) \nabla_{\paramp} \pi(\at \mid \st, \paramp) d\at \, d\st$. \citet{sutton2018reinforcement} show that a generalization of this result includes a comparison of the state-action value function, $q_{\pi}(\st, \at)$, to an arbitrary baseline that does not vary with the action, $\at$. 

When the baseline is the state value function, $v_{\pi}(\st)$, we have an objective in terms of the \emph{advantage function} \citep{baird1993advantage,schulman2015high}, $h_{\pi}(\st, \at) \coloneqq q_{\pi}(\st, \at) - v_{\pi}(\st)$, namely: $\nabla_{\paramp} J(\paramp) \propto \int_{\Scal} \rho(\st) \int_{\Acal} h_{\pi}(\st, \at) \nabla_{\paramp} \pi(\at \mid \st, \paramp) d\at \, d\st$. This formulation in terms of \emph{all actions} can be further simplified in terms of sampled actions and states as $\nabla_{\paramp} J(\paramp) \propto \E_{\pi} \left[ h_{\pi}(\St_{\step}, \At_{\step}) \nabla_{\paramp} \log{\pi(\At_{\step} \mid \St_{\step}, \paramp)} \right]$. We use $\E_{\pi}$ to denote an expectation over states $\St_{\step}$ and actions $\At_{\step}$ collected under the policy $\pi(\at \mid \st)$.

In general, because neither the state-action, $q_{\pi}(\st, \at)$, nor the state value, $v_{\pi}(\st)$, functions are known, we need an estimator for the advantage function. For compactness, we will focus on the generalized advantage estimator (GAE) proposed by \citet{schulman2015high}: $h(\st_{\step}, \rt_{\step}, \paramv) = \sum_{\kstep = \step + 1}^{\Step} (\gamma \lambda)^{\kstep - 1 - \step} \delta_{\step - \kstep + 1}^{\paramv},$ where $0 < \lambda \leq 1$, and $\delta_{\step}^{\paramv} = \rt_{\step} + \gamma v(\st_{\step + 1}; \paramv) - v(\st_{\step}; \paramv)$ is the temporal difference (TD) residual of the value function with discount, $\gamma$ \citep{sutton2018reinforcement}. The GAE then yields a low-variance gradient estimator for the policy function: $\widehat{\nabla_{\paramp} J}(\paramp) \coloneqq \E_{\pi} \left[ h(\St_{\step}, \Rt_{\step}, \paramv) \nabla_{\paramp} \log{\pi(\At_{\step} \mid \St_{\step}, \paramp)} \right]$. Finally, the actor and critic networks are generally optimized by using mini-batch stochastic gradient descent \cite{robbins1951stochastic} to fit the functions induced by the network weights to a batch of data collected under the current policy, $\Dcal^{b}_{\pi} = \{\st_{i}, \at_{i}, \rt_{i}\}_{i=1}^b$. The parameter updates are given by,
\begin{subequations}
    \begin{align}
        \paramp &\gets \paramp - \eta \frac{1}{b} \sum_{i = 1}^{b} h(\st_{i}, \rt_{i}, \paramv) \nabla_{\paramp} \log{\pi(\at_{i} \mid \st_{i}, \paramp)}, \label{eq:actor-update} \\
        \paramv &\gets \paramv - \eta \frac{1}{b} \sum_{i = 1}^{b} \nabla_{\paramv} \log{p(\mathrm{g}(\st_{i}, \rt_{i}, \tilde{\paramv}) \mid \st_{i}, \paramv)},
    \end{align}
    \label{eq:updates}
\end{subequations}
where, $\eta$, is the  learning rate, $\mathrm{g}(\st_{i}, \rt_{i}, \tilde{\paramv}) = h(\st_{i}, \rt_{i}, \tilde{\paramv}) + v(\st_{i}, \tilde{\paramv})$, and $\tilde{\paramv}$ are previous parameters.

\section{Methods}
\label{sec:methods}

This work takes a top-down approach to state-aware exploration for on-policy actor-critic DRL. To employ principled exploration strategies, we would like to have approximate posteriors, $q(\Paramp \mid \Dcal_{n-1})$ and $q(\Paramv \mid \Dcal_{n-1})$, for the weights of the actor and critic given data, $\Dcal_{n-1} = \{\st_{i}, \at_{i}, \rt_{i}\}_{i=1}^{|\Tcal_{n-1}|}$, collected under the policy, $\pi(\at \mid \st, \Paramp_{n-1})$, over a set of horizons, $\Tcal_{n-1} = \Step^{n-1}_1 \cup \Step^{n-1}_2 \cup \dots$. However, fastidiously Bayesian, bottom-up approaches often yield strategies less effective than the state-of-the-art. Leaving debates on evaluation aside, we start from the A3C algorithm and ask, ``what minimum changes can we make to get close to an approximate posterior?"

Starting with the critic, $v(\st, \paramv)$, this task seems straightforward because we train the critic with mean squared error loss. Hence, we can use the implied likelihood, $\mathcal{N}(\mathrm{g} \mid v(\st, \paramv), 1)$, and use the \textit{dropout as a Bayesian approximation} \citep{gal2016dropout} framework to infer an approximate posterior density over critic weights, $q(\paramv \mid \widehat{\paramv}, p_d)$, where $\widehat{\paramv}$ is the variational parameter for the network weights, and $p_d$ is the dropout rate. We outline the resulting optimization procedure in lines 16-17 of \Cref{alg:vsop} for a unit Normal prior over critic weights, $p(\paramv) = \mathcal{N}(\paramv \mid 0, \mathbf{I})$.

\begin{algorithm*}
 \caption{VSOP for Dropout Bayesian Neural Networks \label{alg:vsop}}
 \begin{algorithmic}
 \STATE {\bfseries Input:} initial state, $\st'$, environment, $p(\st', \rt \mid \st, \at)$, rollout buffer, $\Dcal$, initial actor parameters, $\widehat{\paramp}$, initial critic parameters, $\widehat{\paramv}$, dopout rate, $p_d$, learning rate, $\eta$, minibatch size, $b$.
 \REPEAT
 \STATE $\Dcal \gets  \emptyset$ \COMMENT{reset rollout buffer}
 \REPEAT
 \STATE $\st \gets \st'$ \COMMENT{update current state}
 \STATE  $\paramp \sim q(\paramp \mid \widehat{\paramp}, p_d)$ \textbf{if} TS \textbf{else}  $\paramp \gets \widehat{\paramp}$ \COMMENT{sample actor params if Thompson sampling (TS)}
 \STATE $\at \sim \pi(\at \mid \st, \paramp)$ \COMMENT{sample action from policy}
 \STATE $\st', \rt \sim p(\st', \rt \mid \st, \at)$ \COMMENT{sample next state and reward from environment}
 \STATE $\Dcal \gets \Dcal \cup \{(\st, \at, \rt)\}$ \COMMENT{update rollout buffer}
 \UNTIL{rollout ends}
 \STATE $\paramv^* \gets \widehat{\paramv}$ \COMMENT{freeze critic weights for advantage estimates}
 \STATE $\beta \gets (1 - p_d) / \left(2|\Dcal|\right)$ \COMMENT{set parameter precision}
 \REPEAT
 \STATE $\{\st_i, \at_i, \rt_i\}_{i=1}^{b} \sim \Dcal$ \COMMENT{sample minibatch from rollout buffer}
 \STATE  $\widetilde{\paramv} \sim q(\paramv \mid \paramv^*, p)$ \textbf{if} TS \textbf{else}  $\widetilde{\paramv} \gets \paramv^*$ \COMMENT{sample advantage params if TS}
 \STATE $\paramp \sim q(\paramp \mid \widehat{\paramp}, p_d)$ \COMMENT{sample actor parameters}
 \STATE $\widehat{\paramp}\gets \widehat{\paramp} - \eta \frac{1}{b} \sum_{i = 1}^{b} h^+(\st_{i}, \rt_{i}, \widetilde{\paramv}) \nabla_{\paramp} \log{\pi(\at_{i} \mid \st_{i}, \paramp)} + 2\beta\paramp$ \COMMENT{update actor}
 \STATE $\paramv \sim q(\paramv \mid \widehat{\paramv}, p_d)$ \COMMENT{sample critic parameters}
 \STATE $\widehat{\paramv} \gets \widehat{\paramv} - \eta \frac{1}{b} \sum_{i = 1}^{b} \nabla_{\paramv} \log{p(\mathrm{g}(\st_{i}, \rt_{i}, \widetilde{\paramv}) \mid \st_{i}, \paramv)} + 2\beta\paramv$ \COMMENT{update critic}
 \UNTIL{actor and critic update complete}
 \UNTIL{finished}
 \end{algorithmic}
\end{algorithm*}

The inference task is less straightforward for the actor because the A3C objective, $\nabla_{\paramp} J(\paramp) \propto \E_{\pi} \left[ h_{\pi}(\St_{\step}, \At_{\step}) \nabla_{\paramp} \log{\pi(\At_{\step} \mid \St_{\step}, \paramp)} \right]$, is not merely maximization of the log-likelihood. Instead, the log-likelihoods, $\log{\pi(\at \mid \st, \paramp)}$, are scaled by the advantage function, $h_{\pi}(\st, \at)$. The \textit{dropout as Bayesian approximation} framework estimates the integral over the log evidence lower bound objective using Monte-Carlo integration. For a single sample from the approximate posterior density, $\paramp \sim q(\paramp \mid \widehat{\paramp}, p_d)$, the integrand is of the general form:
\begin{equation}
 \mathcal{L} = \sum_{i=1}^{|\mathcal{D}|} \log p(\cdot \mid \st_{i}, \paramp) - \mathrm{KL}(q(\paramp) || p(\paramp)).
 \label{eq:elbo_integrand}
\end{equation}
We make the same prior assumption, $p(\paramp) = \mathcal{N}(\paramp \mid 0, \mathbf{I})$, for the actor as for the critic, so we only need to focus on the first r.h.s., log-likelihood term. Here we introduce a normal-gamma distribution over the actions r.v., $\At_{\step}$, and a random variable $\Ht_{\step}$:
\begin{equation*}
 \begin{split}
 p(\At_{\step}, &\Ht_{\step} \mid \st, \paramp, \tau, \alpha, \beta)  \\
 &\coloneqq\begin{cases}
 \At_{\step} \mid \Ht_{\step}, \st, \paramp, \tau \sim \mathcal{N}\left(\at \mid \mu(\st, \paramp), (\tau \Ht_{\step})^{-1}\right) \\
 \Ht_{\step} \mid \alpha, \beta \sim \text{Gamma}(\alpha, \beta).
 \end{cases}
 \end{split}
\end{equation*}
Given a dataset, $\mathcal{D} = \{\At_i, \St_i, \Ht_i\}_{i=1}^N$, and differentiating the log-likelihood with respect to $\paramp$, we have:
\begin{equation}
 \begin{split}
 \nabla&_{\paramp}\log{p(\At_{\step}, \Ht_{\step} \mid \St_{\step}, \paramp, \dots)} \\
 &= \sum_{i=1}^N\nabla_{\paramp}\log\Bigg(
 \frac{\beta^{\alpha}\sqrt{\tau}}{\Gamma(\alpha)\sqrt{2\pi}}
 \mathrm{h}_i^{\alpha - \frac{1}{2}}
 \exp\left(-\beta\mathrm{h}_i\right) \\
 &\quad\quad\quad\quad\quad\quad\exp\left(-\frac{1}{2}\mathrm{h}_i\tau(\at_i - \mu(\st_i, \paramp))^2\right)
 \Bigg) \\
 &= -\frac{1}{2} \sum_{i=1}^{N}\mathrm{h}_i\nabla_{\paramp}\tau \left(\at_i - \mu(\st_i, \paramp)\right)^2 \\
 &= -\frac{1}{2} \sum_{i=1}^{N}\mathrm{h}_i\nabla_{\paramp}\log{\pi(\at_{\step} \mid \st_{\step}, \paramp)}.
 \end{split}
 \label{eq:gamma}
\end{equation}
Thus, the normal-gamma assumption allows us to recover the form of the A3C gradient estimator in \Cref{eq:actor-update} while enabling approximate Bayesian inference over the actor parameters. However, the advantage function, $h_{\pi}(\st, \at)$, is not gamma distributed, as it can take on values in the range $(-\infty, 0]$, so we cannot plug it directly into \Cref{eq:gamma}. This discrepancy motivates our second, though not strictly valid due to the introduction of zeros, minimal step of passing the advantages through a ReLU function, enabling approximate Bayesian inference over the parameters of the actor to obtain the approximate posterior of the actor parameters, $q(\paramp \mid \widehat{\paramp}, p_d)$. 

Sampling a policy from this approximate posterior involves sampling a dropout mask and running a forward pass of the network, yielding the policy, $\pi(\mathbf{a} \mid \mathbf{s}, \paramp)$. Then, sampling an action is done by sampling an action from the sampled policy, $\mathbf{a} \sim \pi(\mathbf{a} \mid \mathbf{s}, \paramp)$. We outline this Thompson sampling \citep{thompson1933likelihood} procedure in lines 5-6 of \Cref{alg:vsop}.

We hypothesize that this is a better state-aware exploration method for two reasons. First, it is adaptive: for less frequently visited states the diversity of the sampled parameters of the policy will be greater, promoting more exploration. As a state is visited more often under actions that yield positive advantages, the diversity of samples will concentrate, promoting less exploration. Thus, we get more exploration for states where the actor has less experience of good actions and less exploration in states where the actor's decisions have led to good expected returns. Second, this exploration is done around the mode of the policy distribution, resulting in more conservative exploration.

We outline an optimization step of the actor in lines 14-15 of \Cref{alg:vsop}, where $h^+(\st_{i}, \rt_{i}, \widetilde{\paramv}) \coloneqq \max\big(0, h(\st_{i}, \rt_{i}, \widetilde{\paramv})\big)$.
Note that the clipped advantages as an actor precision modifier have a very intuitive interpretation.
When advantage estimates are low (no evidence of past good actions), the variance of the policy will be high, indicating that the actor should explore more.
Conversely, when the advantage estimates are high (evidence of past good actions), the variance of the policy will be low, indicating that the actor should explore less.
We provide further commentary in \Cref{app:commentary-gamma}.

\textbf{What Function Does VSOP Optimize?}
Clipping the advantage estimates to be non-negative has been explored in many policy-gradient contexts \citep{srinivasan2018actor,oh2018self,petersen2019deep,ferret2020self}. Here, we examine how this augmentation affects the optimization procedure in the context of on-policy actor-critic RL and offer a theoretical hypothesis to ground both our method and the Regret Matching Policy Gradient (RMPG) method of \citet{srinivasan2018actor}.

\begin{theorem}
 \label{th:main}
 Let, $\Gt_{\step} \coloneqq \sum_{\kstep = \step + 1}^{\Step} \gamma^{\kstep - 1 - \step}  \Rt_{\kstep}$, denote the discounted return.
 Let $q_{\pi}(\st, \at) = \E_{\pi} \left[ \Gt_{\step} \mid \St_{\step} = \st,  \At_{\step} = \at\right]$, denote the state-action value function, and
 $v_{\pi}(\st) = \E_{\pi} \left[ \Gt_{\step} \mid \St_{\step} = \st \right]$, denote the state value function, under policy $\pi(\at \mid \st, \paramp)$.
 Define the ReLU function as $\big(x\big)^+ \coloneqq \max(0, x)$.
 Assume that rewards, $\Rt_{\step}$, are non-negative and the gradient of the policy, $\nabla \pi(\at \mid \st, \paramp)$, is a conservative vector field.
 Then, performing gradient ascent with respect to,
 \begin{equation*}
 \begin{split}
 \nabla_{\paramp}& J(\paramp) \coloneqq  \\
 &\E_{\pi}\left[ \Bigl(q_{\pi}(\St_{\step}, \At_{\step}) - v_{\pi}(\St_{\step}) \Bigr)^+ \nabla_{\paramp} \log \pi(\At_{\step} \mid \St_{\step}, \paramp) \right],
 \end{split}
 \label{eq:main-objective}
 \end{equation*}
 maximizes a lower-bound, $v_{\pi}^*(\st)$, on the state value function, $v_{\pi}(\st)$, plus an additive term:
 \begin{equation}
 v_{\pi}^*(\st) \leq v_{\pi}(\st) + C_{\pi}(\st).
 \label{eq:main-bound}
 \end{equation}
 Where, $C_{\pi}(\st) = \iint \Big( \gamma v_{\pi}(\st') - v_{\pi}(\st) \Big)^{+} d\Prob(\st' \mid \St_{\step}=\st, \At_{\step} = \at) d\Pi(\at \mid \St_{\step}=\st)$, is the expected, clipped difference in the state value function, $\gamma v_{\pi}(\st') - v_{\pi}(\st)$, over all actions, $\at$, and next states, $\st'$, under the policy given state, $\st$. Here, we use $\int \dots d\Pi(\at \mid \st)$ to denote $\sum_{\at} \dots \pi(\at \mid \st)$ for discrete action spaces and $\int \dots \pi(\at \mid \st)d\at$ for continuous action spaces. Similarly, we use $\int \dots d\Prob(\st' \mid \st, \at)$ to denote $\sum_{\st'} \dots p(\st' \mid \st, \at)$ for discrete state spaces and $\int \dots p(\st' \mid \st, \at)d\st'$ for continuous state spaces. We provide the proof in \Cref{app:constant-proof}.
\end{theorem}

\textbf{Bounding $C_\pi(\st)$.}
For a $K_{\pi}$-Lipschitz value function and $\gamma=1$, the additive term is bounded proportional to the expected absolute difference between states:
\begin{subequations}
 \begin{align*}
 C_{\pi}(\st)
 &= \iint \Big( v_{\pi}(\st') - v_{\pi}(\st) \Big)^{+} d\Prob(\st' \mid \st, \at) d\Pi(\at \mid \st) \\
 &\leq= \frac{1}{2} \iint \big| v_{\pi}(\st') - v_{\pi}(\st) \big| d\Prob(\st' \mid \st, \at) d\Pi(\at \mid \st) \\
 &\leq \frac{1}{2} \iint K_{\pi}\big|\big| \st' - \st \big|\big| d\Prob(\st' \mid \st, \at) d\Pi(\at \mid \st),
 \end{align*}
\end{subequations}
where the second line follows from $\Cref{lem:relu-lt-abs}$.
This interpretation motivates using spectral normalization \citep{miyato2018spectral} of the value function estimator weights, $v(\st, \paramv)$, which regulates the Lipschitz constant, $K_{\pi}$, of the estimator and can improve off-policy DRL performance \citep{bjorck2021towards,gogianu2021spectral}. Moreover, this bound is not vacuous for the continuous (nor the discrete) action setting. Under weak assumptions, $f(\at, \st) \coloneqq \int K_{\pi}\big|\big| \st' - \st \big|\big| d\Prob(\st' \mid \St_{\step}=\st, \At_{\step} = \at)$, is finite for all $\at$. Therefore, $f^*(\st) = \max_{\at}(\int f(\at, \st)d\Pi(\at \mid \St_{\step}=\st))$, exists and is finite, and $C_{\pi}(\st) \leq \frac{1}{2}f^*(\st)$.
We provide further commentary concerning the constant $C_{\pi}(\st)$ and Lipschitz continuity in \Cref{app:commentary-lipschitz}.

We term this method VSOP for Variational [b]ayes, Spectral-normalized, On-Policy reinforcement learning. \Cref{alg:vsop} details VSOP for dropout BNNs.

\section{Related Works}
\label{sec:related-works}
VSOP is an on-policy RL algorithm. \Cref{table:performance_metrics_full} in \Cref{app:related-works} compares the gradient of the performance function, $\nabla J(\paramp)$, for VSOP with those for relevant on-policy algorithms.

\textbf{Maximum Entropy RL.}
\citet{levine2018reinforcement} establishes the theoretical connection between maximum entropy RL and parameter agnostic probabilistic inference of policies over optimal trajectories. Further, in Section 4.1 of that work, he shows how approximate inference in maximum entropy policy gradients setting corresponds precisely to the gradient estimator of the advantage actor-critic algorithm (A2C or A3C). It is clear from \Cref{th:main} that our gradient estimator is distinct from the advantage actor-critic estimator and that we optimize a fundamentally different objective than the maximum entropy objective. Therefore, we maintain that our step towards approximate inference over policy parameters represents a distinct alternative to the maximum entropy paradigm.

\textbf{Mirror Learning.}
\emph{Proximal Policy Optimization (PPO)} \citep{schulman2017proximal}, improves upon the baseline policy gradient method by constraining the maximum size of policy updates. PPO employs a clipped surrogate objective function to limit the size of policy updates. PPO simplifies the optimization procedure compared to TRPO \citep{schulman2015trust}, making it more computationally efficient and easier to implement. While PPO constrains policy updates based on the ratio between the new and old policies, VSOP constrains policy updates according to the sign of the estimated advantage function. As such, PPO is an instance of the \emph{mirror learning} framework \cite{kuba2022mirror}, whereas VSOP does not inherit the same theoretical guarantees.
\citet{lu2022discovered} explores the Mirror Learning space by meta-learning a “drift” function. They term their immediate result Learned Policy Optimization (LPO). Through its analysis, they arrive at \emph{Discovered Policy Optimisation (DPO)}, a novel, closed-form RL algorithm.

\textbf{Regret Matching Policy Gradient (RMPG).} \citet{srinivasan2018actor} present a method inspired by their regret policy gradient (RPG) objective, which maximizes a lower-bound on the advantages: $(h(\st, \at))^{+} \leq h(\st, \at)$. RPG directly optimizes the policy for an estimator of the advantage lower-bound, denoted as $\nabla_{\paramp} J^{\textrm{RPG}}(\paramp)$. RMPG, being inspired by RPG, has a different objective, $\nabla_{\paramp} J^{\textrm{RMPG}}(\paramp)$. In both cases, $q(\st, \at, \paramv)$ is a parametric estimator of the state-action value function, $q_{\pi}(\st, \at)$.  RMPG has demonstrated improved sample efficiency and stability in learning compared to standard policy gradient methods.
VSOP is closely related to RMPG; however, we provide the missing theoretical foundations to ground RMPG (\Cref{app:constant-proof}), extend RMPG from the \emph{all actions} formulation making it more suitable for continuous control (\Cref{sec:rmpg-to-vsop}), and employ the GAE rather than the state-action value function estimator, $q(\st, \at, \paramv)$.

\begin{figure*}[ht]
    \centering
    \begin{subfigure}[b]{0.72\textwidth}
        \centering
        \includegraphics[width=\textwidth]{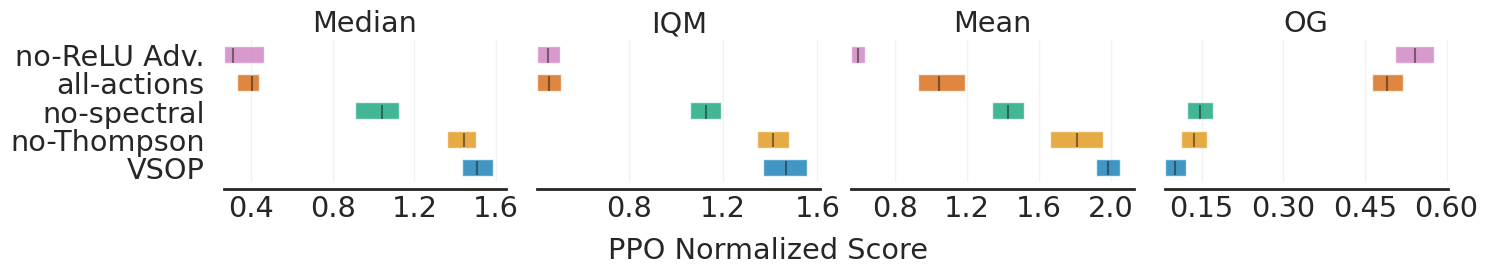}
        \caption{Episodic return}
        \label{fig:sprectral-performance}
    \end{subfigure}
    ~
    \begin{subfigure}[b]{0.26\textwidth}
        \centering
        \includegraphics[width=\textwidth]{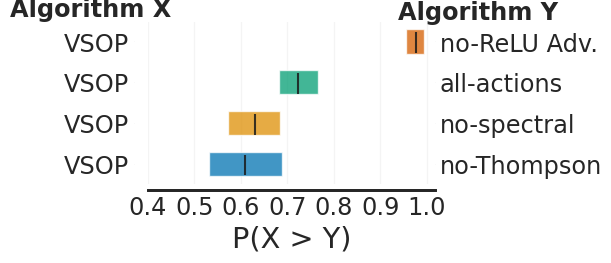}
        \caption{Prob. Improve}
        \label{fig:sprectral-performance-pi}
    \end{subfigure}
    
    \begin{subfigure}[b]{0.72\textwidth}
        \centering
        \includegraphics[width=\textwidth]{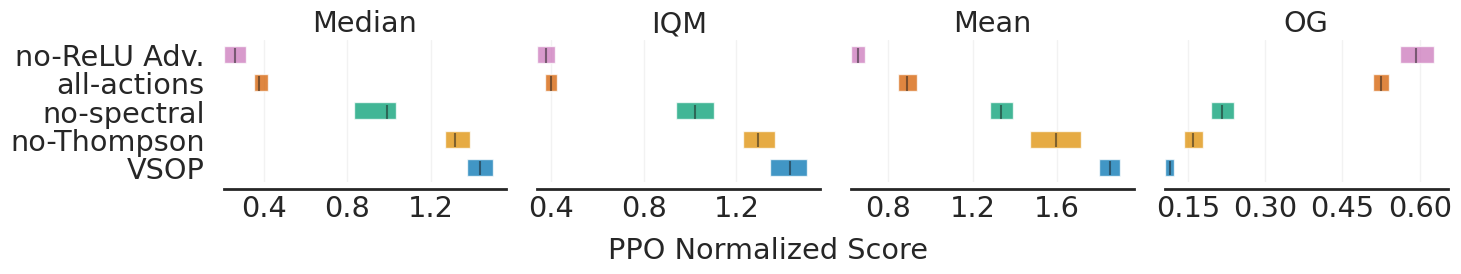}
        \caption{Area Under Return Curve}
        \label{fig:sprectral-efficiency}
    \end{subfigure}
    ~
    \begin{subfigure}[b]{0.26\textwidth}
        \centering
        \includegraphics[width=\textwidth]{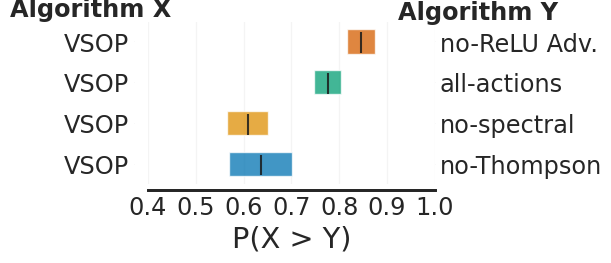}
        \caption{Prob. Improve}
        \label{fig:sprectral-efficiency-pi}
    \end{subfigure}
    \caption{MuJoCo. Ablating the effect of the proposed mechanisms. Here, we compare VSOP to VSOP without spectral normalization (no-spectral), VSOP without Thompson sampling (no-Thompson), VSOP without advantage clipping (no-ReLU Adv.), and VSOP using all-actions policy optimization (all actions). We see that no single mechanism contributes greater than the sum of all changes, lending credence to the validity of our theory. The overall performance (a-b) and sample efficiency (c-d) metrics illustrate this result. Metrics are computed wrt to the average episodic return of the last 100 episodes and the area under the episodic return curve over ten random seeds}
    \label{fig:rliable-mechanisms}
\end{figure*}

\textbf{Thompson Sampling in Deep Reinforcement Learning.}
Thompson sampling has been extensively explored in conventional and Deep Q-Learning \citep{strens2000bayesian,wang2005bayesian,osband2016deep,moerland2017efficient,azizzadenesheli2018efficient} to improve exploration and sample efficiency.
\citet{clements2019estimating} and \citet{nikolov2018information} propose similar sampling-based exploration strategies for Deep Q-Learning.
\citet{jiang2023importance} propose a Thompson sampling strategy based on an ensemble of quantile estimators of the state-action value distribution.
In the context of \emph{policy gradient} methods, related Upper Confidence Bound (UCB) \citep{ciosek2019better} and Hamiltonian Monte-Carlo (HMC) \citep{xu2022improving} approaches are proposed for off-policy Soft Actor-Critic (SAC) \citep{haarnoja2018soft}, and \citet{henaff2022exploration} proposes an elliptical episodic reward for general use. 
\citet{hausknecht2022consistent} and \citet{igl2019generalization} use fixed dropout masks to sample policies and actions but stopped short of formalizing this as Thompson sampling.
Our work formalizes and shows the benefit of Thompson sampling for on-policy actor-critic methods.

\section{Experiments}
We evaluate VSOP against on-policy RL methods across various domains, including continuous and discrete action spaces and diverse dimensionalities in both the action and observation spaces. 
In \Cref{sec:exp-gymnasium}, we evaluate VSOP on continuous control tasks using the Gymnasium \citep{brockman2016openai} and Gymnax \citep{gymnax2022github} implementations of MuJoCo \citep{todorov2012mujoco}.
In \Cref{sec:exp-procgen}, we assess the capacity of VSOP to learn policies that generalize to unseen environments at test time using the ProcGen benchmark \citep{cobbe2020leveraging}.
We use the rliable package to evaluate robust normalized median (Median), interquartile mean (IQM), mean (Mean), optimality gap (OG), and probability of improvement (Prob. Improve) metrics \cite{agarwal2021deep}.
We provide additional results in \Cref{app:additional-results} and make code available at \href{https://github.com/anndvision/vsop}{https://github.com/anndvision/vsop}.

\subsection{MuJoCo}
\label{sec:exp-gymnasium}
For this evaluation, we build off of \citet{huang2022cleanrl}'s \href{https://github.com/vwxyzjn/cleanrl}{CleanRL} package which provides reproducible, user-friendly implementations of state-of-the-art reinforcement learning algorithms using PyTorch \citep{paszke2019pytorch}, Gymnasium \citep{brockman2016openai,todorov2012mujoco}, and Weights \& \citep{wandb}.
We give full implementation details in \Cref{app:implementation-details-gym}.

\begin{figure*}[ht]
    \centering
    \begin{subfigure}[b]{0.195\textwidth}
        \centering
        \includegraphics[width=\textwidth]{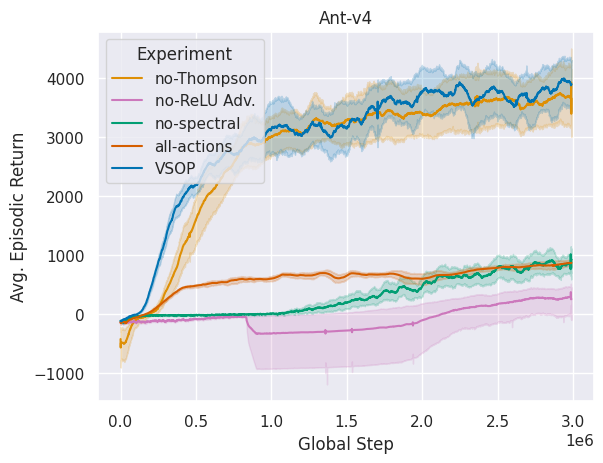}
        \caption{Ant}
        \label{fig:mujoco-ant}
    \end{subfigure}
    \hfill
    \begin{subfigure}[b]{0.195\textwidth}
        \centering
        \includegraphics[width=\textwidth]{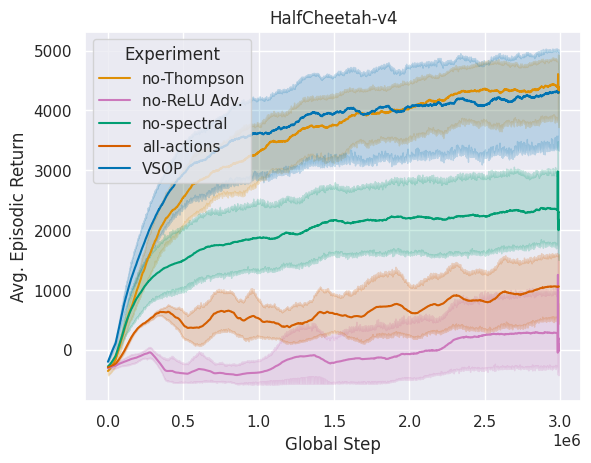}
        \caption{HalfCheetah}
        \label{fig:mujoco-halfcheetah}
    \end{subfigure}
    \hfill
    \begin{subfigure}[b]{0.195\textwidth}
        \centering
        \includegraphics[width=\textwidth]{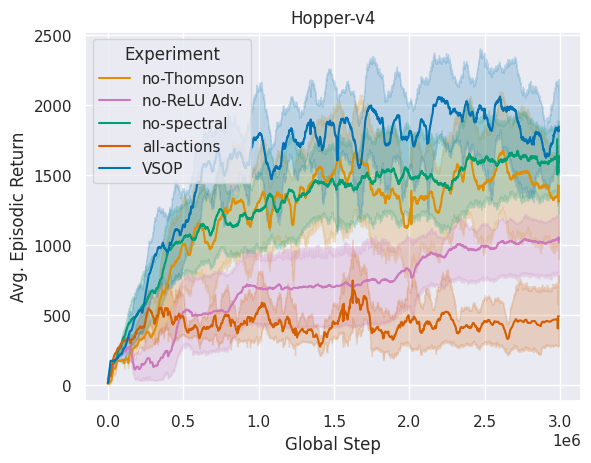}
        \caption{Hopper}
        \label{fig:mujoco-hopper}
    \end{subfigure}
    \hfill
    \begin{subfigure}[b]{0.195\textwidth}
        \centering
        \includegraphics[width=\textwidth]{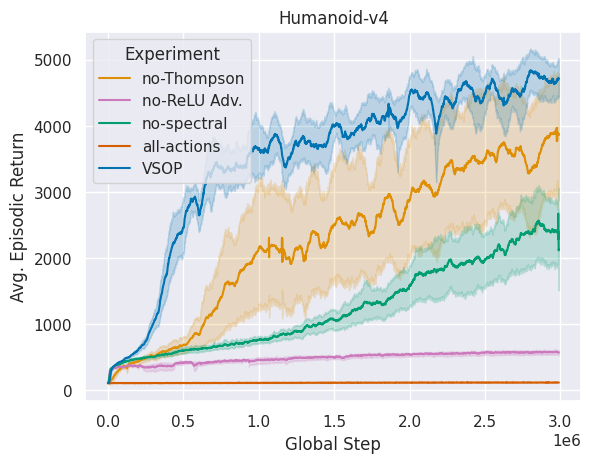}
        \caption{Humanoid}
        \label{fig:mujoco-humanoid}
    \end{subfigure}
    \hfill
    \begin{subfigure}[b]{0.195\textwidth}
        \centering
        \includegraphics[width=\textwidth]{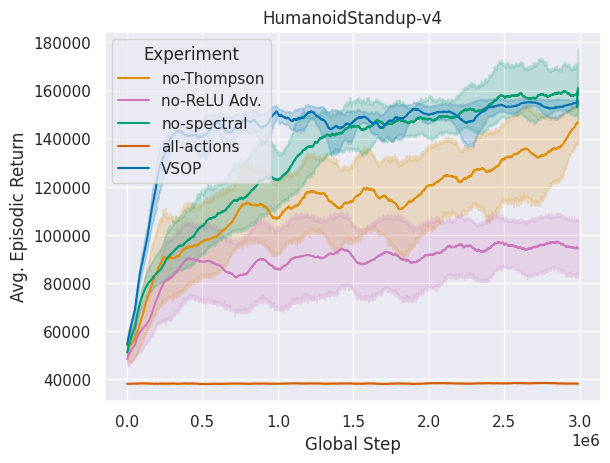}
        \caption{HumanoidStandup}
        \label{fig:mujoco-humanoidstandup}
    \end{subfigure}
    
    \begin{subfigure}[b]{0.195\textwidth}
        \centering
        \includegraphics[width=\textwidth]{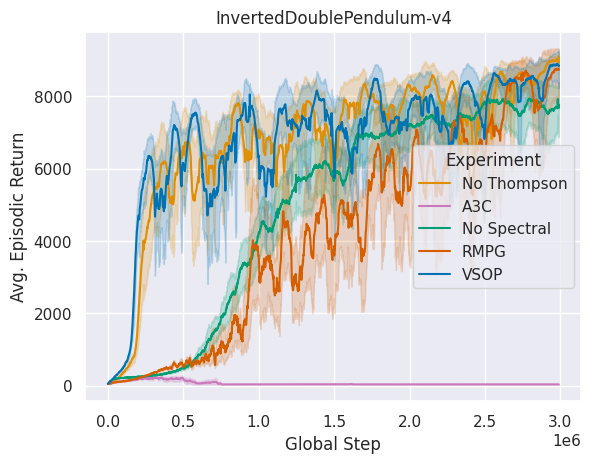}
        \caption{DoublePendulum}
        \label{fig:mujoco-inverted_double_pendulum}
    \end{subfigure}
    \hfill
    \begin{subfigure}[b]{0.195\textwidth}
        \centering
        \includegraphics[width=\textwidth]{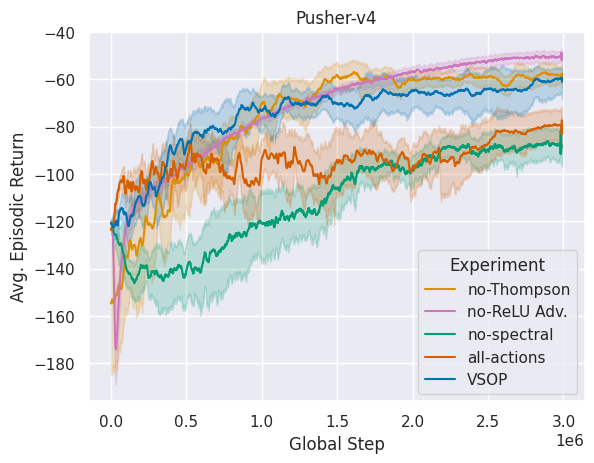}
        \caption{Pusher}
        \label{fig:mujoco-pusher}
    \end{subfigure}
    \hfill
    \begin{subfigure}[b]{0.195\textwidth}
        \centering
        \includegraphics[width=\textwidth]{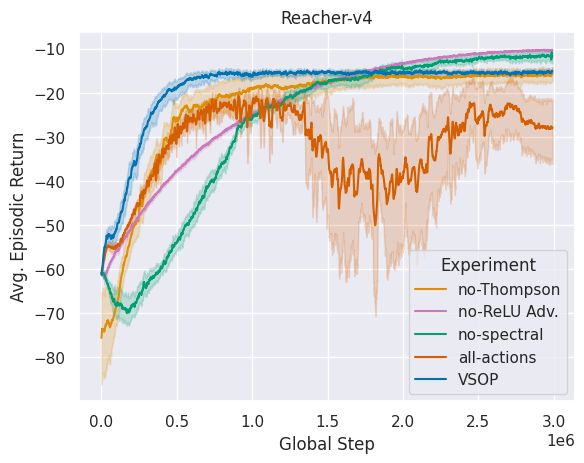}
        \caption{Reacher}
        \label{fig:mujoco-reacher}
    \end{subfigure}
    \hfill
    \begin{subfigure}[b]{0.195\textwidth}
        \centering
        \includegraphics[width=\textwidth]{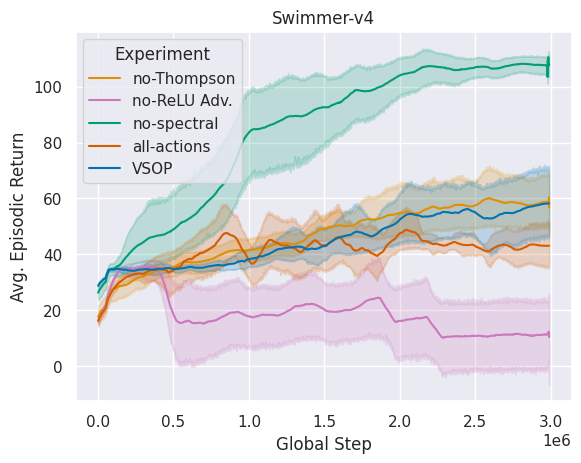}
        \caption{Swimmer}
        \label{fig:mujoco-swimmer}
    \end{subfigure}
    \hfill
    \begin{subfigure}[b]{0.195\textwidth}
        \centering
        \includegraphics[width=\textwidth]{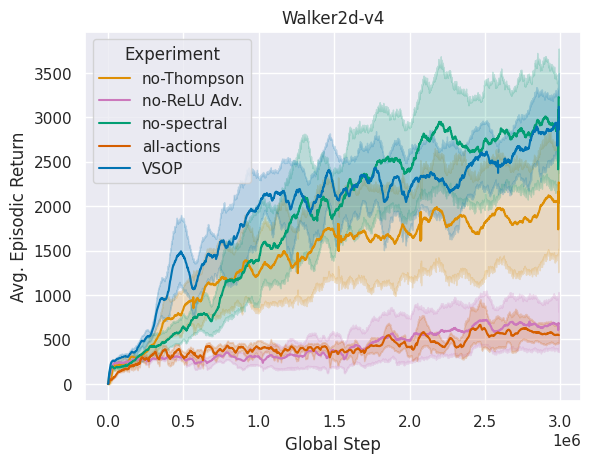}
        \caption{Walker2d}
        \label{fig:mujoco-walker2d}
    \end{subfigure}
    \caption{
        Comparing the effect of VSOP mechanisms on Mujoco continuous control performance.
        Using the single action framework and updating the policy only on positive advantage estimates have the largest effects, followed by spectral normalization, and finally Thompson sampling.
        Blue lines (VSOP) show the optimized proposed method. 
        Orange lines (no-Thompson) show VSOP without Thompson sampling. 
        Green lines (no-Spectral) show VSOP without spectral normalization.
        Pink lines (all actions) show VSOP with ``all actions''.
        Red lines (no ReLU Adv.) show VSOP without restricting policy updates to positive advantages.
    }
    \label{fig:mujoco-hyper}
\end{figure*}

\textbf{Ablation of Mechanisms.}
First, we investigate the influence of our four proposed mechanisms on the performance of VSOP.
For reference, the mechanisms are positive advantages, single-action setting, spectral normalization, and Thompson sampling.
To ablate each mechanism, we compare VSOP to four variants: VSOP without advantage clipping (no-ReLU Adv.), VSOP in the all-actions setting (all-actions), VSOP without spectral normalization (no-spectral), and VSOP without Thompson sampling (no-Thompson).
We hyperparameter tune each variant by the same procedure used for VSOP (see \Cref{table:gymnasium-ablation-hypers} for details).
\Cref{fig:rliable-mechanisms} summarizes these results, and we see clearly that no single mechanism contributes greater than the sum of all changes, lending credence to our theoretical analysis. 
We see that positive advantages and operating in the single-action regime impact performance on MuJoCo significantly.
Spectral normalization and Thompson sampling also influence performance on MuJoCo positively, especially in high-dimensional action and observation space settings such as Humanoid, Humanoid Stand-Up, and Ant, as shown in \Cref{fig:mujoco-hyper}.
The performance gains for spectral normalization align with results given by \citet{bjorck2021towards} and \citet{gogianu2021spectral} for DDPG \citep{lillicrap2015continuous}, DRQ \citep{kostrikov2020image}, Dreamer \citep{hafner2019dream}, DQN \citep{mnih2015human,wang2016dueling} and C51 \citep{bellemare2017distributional}.

\begin{figure}[ht]
    \centering
    \includegraphics[width=\linewidth]{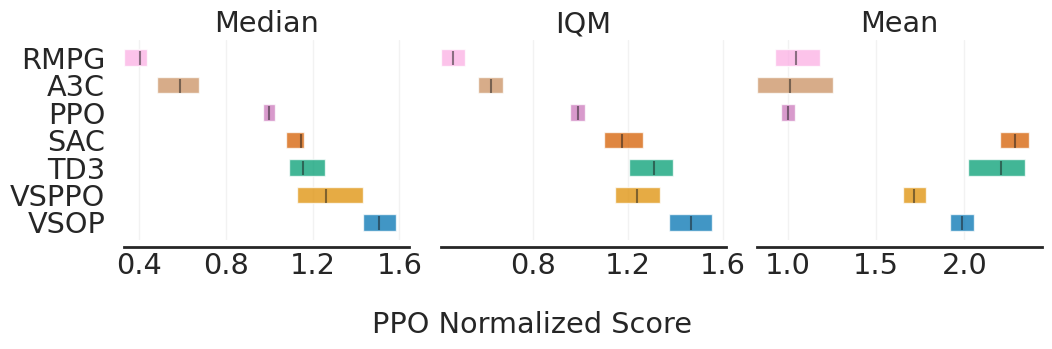}
    \caption{MuJoCo. Comparison to baselines. We see that VSOP (blue) shows significant improvement over each baseline concerning the Median and IQM metrics. VSOP only trails SAC and TD3 for the mean metric. Metrics are computed wrt to the average episodic return of the last 100 episodes over 10 random seeds}
    \label{fig:rliable-baselines}
\end{figure}

\textbf{Comparison to Baselines.}
Next, we compare VSOP to baseline implementations: PPO, A3C, SAC, and TD3. 
We use the CleanRL \citep{huang2022cleanrl} implementation of PPO, SAC, and TD3; the StableBaselines3 \citep{stable-baselines3} hyper-parameter settings for A3C.
We also include comparisons to RMPG (adapted for continuous control) and VSPPO (PPO with spectral normalization, and Thompson sampling via dropout).
We tune RMPG and VSPPO using the same Bayesian optimization \citep{snoek2012practical} protocol as VSOP.
\Cref{fig:rliable-baselines} summarizes our results, where we see that VSOP shows significant improvement over each baseline concerning each metric, except for the SAC and TD3's mean scores.
See \Cref{fig:mujoco-baselines} in \Cref{app:additional-results-baselines} for training curves of these results. 

\begin{figure}[ht]
    \centering
    \includegraphics[width=\linewidth]{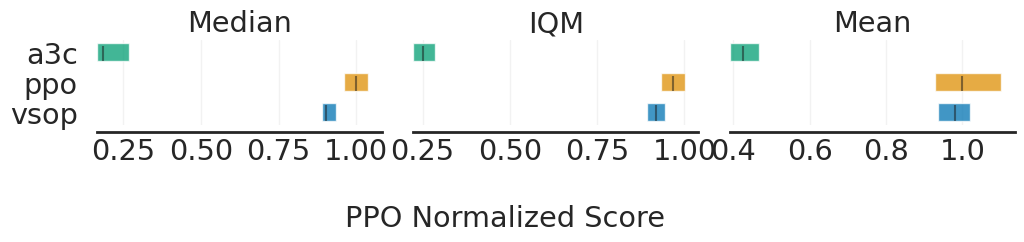}
    \caption{MuJoCo. Comparison to on-policy baselines with extreme parallelization. We compare VSOP to on-policy baselines on MuJoCo with 2048 threads and 10 steps per rollout. Metrics are computed wrt to the average episodic return of the last 100 episodes over 20 random seeds}
    \label{fig:rliable-brax-mujoco}
\end{figure}

\textbf{Effect of Asynchronous Parallelization.}
Following \citet{lu2022discovered}, we build on PureJaxRL to evaluate VSOP on the Brax implementation of MuJoCo in a massively parallel setting. 
Where in the above experiments we set the number of asynchronous threads to 1 and the number of steps per rollout to 2048, here we set the number of asynchronous threads to 2048 and the number of steps to 10.
We see in \Cref{fig:rliable-brax-mujoco} that while VSOP still outperforms A3C significantly, it trails PPO.
Full training curves are shown in \Cref{fig:brax} of \Cref{sec:exp-gymnax}.

\begin{figure*}[ht]
    \centering
    \begin{subfigure}[b]{0.49\textwidth}
        \centering
        \includegraphics[width=\textwidth]{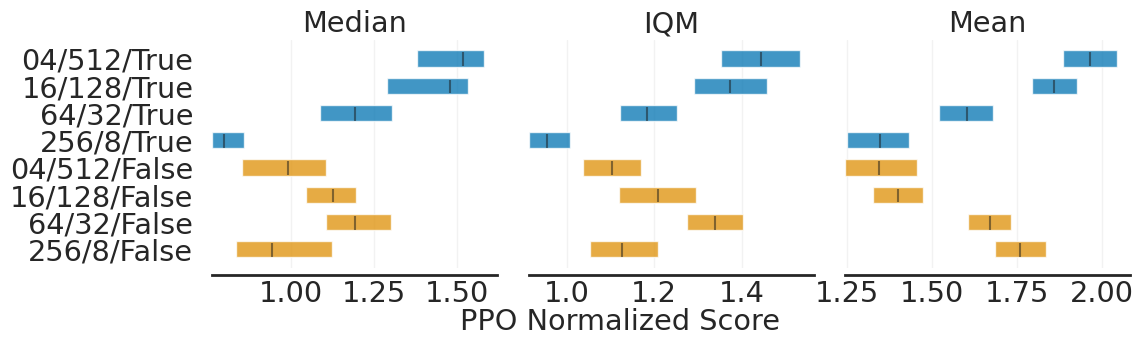}
        \caption{Episodic return}
        \label{fig:spectral-performance}
    \end{subfigure}
    \begin{subfigure}[b]{0.49\textwidth}
        \centering
        \includegraphics[width=\textwidth]{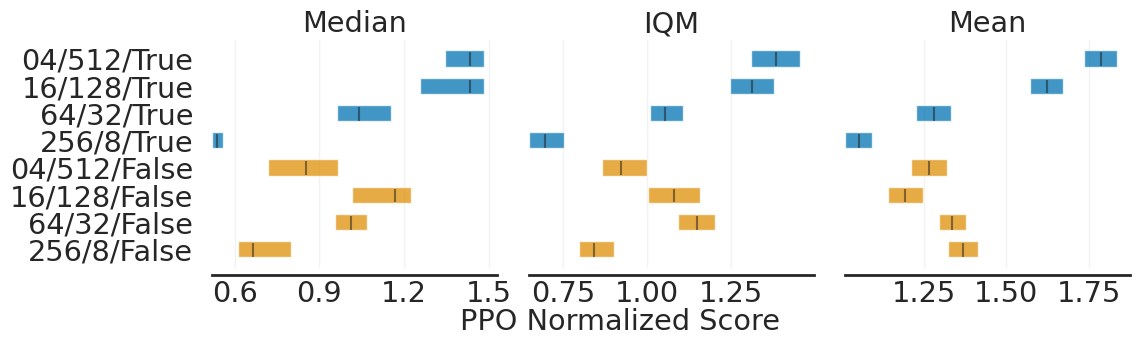}
        \caption{Area under the curve}
        \label{fig:spectral-efficiency}
    \end{subfigure}
    \caption{MuJoCo: effect of parallelization on VSOP. Naming convention: \#threads/\#steps/spectral norm. We see that VSOP is most effective \ref{fig:spectral-performance} and most efficient \ref{fig:spectral-efficiency} in lower thread settings for a fixed rollout size of 2048 steps when using spectral normalization. Metrics are computed wrt to the average episodic return or area under the curve for the last 100 episodes over 5 random seeds}
    \label{fig:rliable-specrtal}
\end{figure*}

\begin{figure*}[ht]
    \centering
    \begin{subfigure}[b]{\textwidth}
        \centering
        \includegraphics[width=0.6\textwidth]{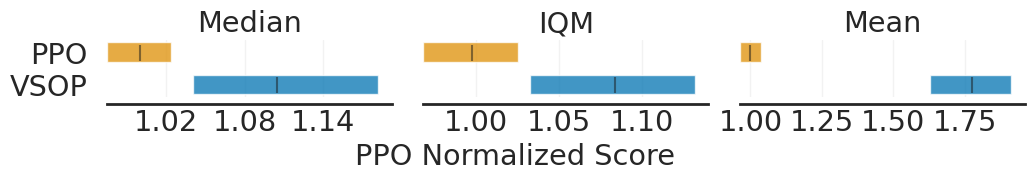}
    \end{subfigure}
    \begin{subfigure}[b]{\textwidth}
        \centering
        \includegraphics[width=0.8\textwidth]{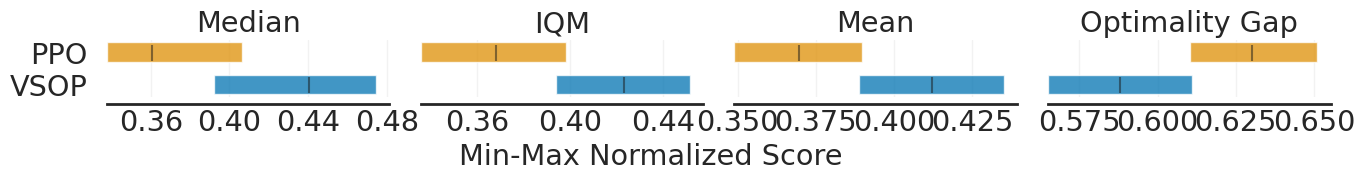}
    \end{subfigure}
    \begin{subfigure}[b]{\textwidth}
        \centering
        \includegraphics[width=0.8\textwidth]{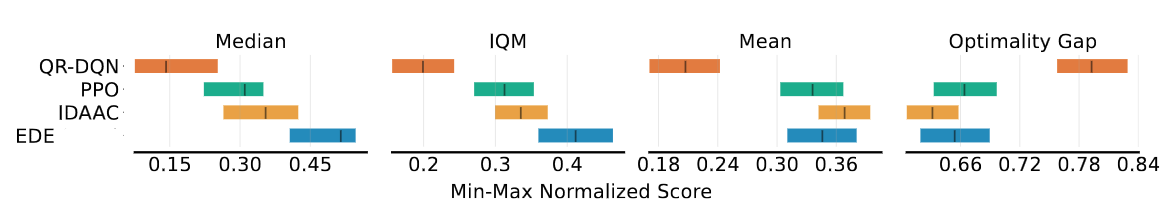}
    \end{subfigure}
    \caption{ProcGen comparison to PPO. In the top pane, we see significant improvement over PPO concerning all metrics for the PPO normalized scores. In the middle pane, we see significant improvement over PPO in terms of the IQM, mean, and optimality gap metrics for the Min-Max normalized scores. In the bottom pane, we include results reported by \citet{jiang2023importance}. It appears as though we improved over EDE with respect to the IQM, mean, and optimality gap metrics. Metrics are computed wrt to the average episodic return of the last 100 episodes over 5 random seeds}
    \label{fig:rliable-procgen}
    
\end{figure*}

Interestingly, hyper-parameter tuning showed spectral normalization to be detrimental to the performance of VSOP in this massively parallel setting.
We investigate the effect of parallelization on VSOP effectiveness and efficiency in \Cref{fig:rliable-specrtal}.
Here we set the rollout size to 2048 environment interactions and sweep the number of threads and number of steps.
For each configuration, we do a hyper-parameter sweep in MuJoCo Brax using the reacher, hopper, and humanoid environments over 1 million environment interactions. 
We then evaluate 10 MuJoCo environments over 3 million environment interactions.
The blue bars show metrics for VSOP with spectral normalization. 
We see that VSOP is most effective and efficient with spectral normalization with a low thread count and that for a fixed rollout size, these measures fall with increasing parallelization.
For VSOP without spectral normalization, the trend is less clear but appears to be generally the opposite for a fixed rollout size.
This indicates that spectral normalization will be beneficial in applications where it is not feasible to run many parallel agents.

\begin{figure*}[ht]
    \centering
    \includegraphics[width=\textwidth]{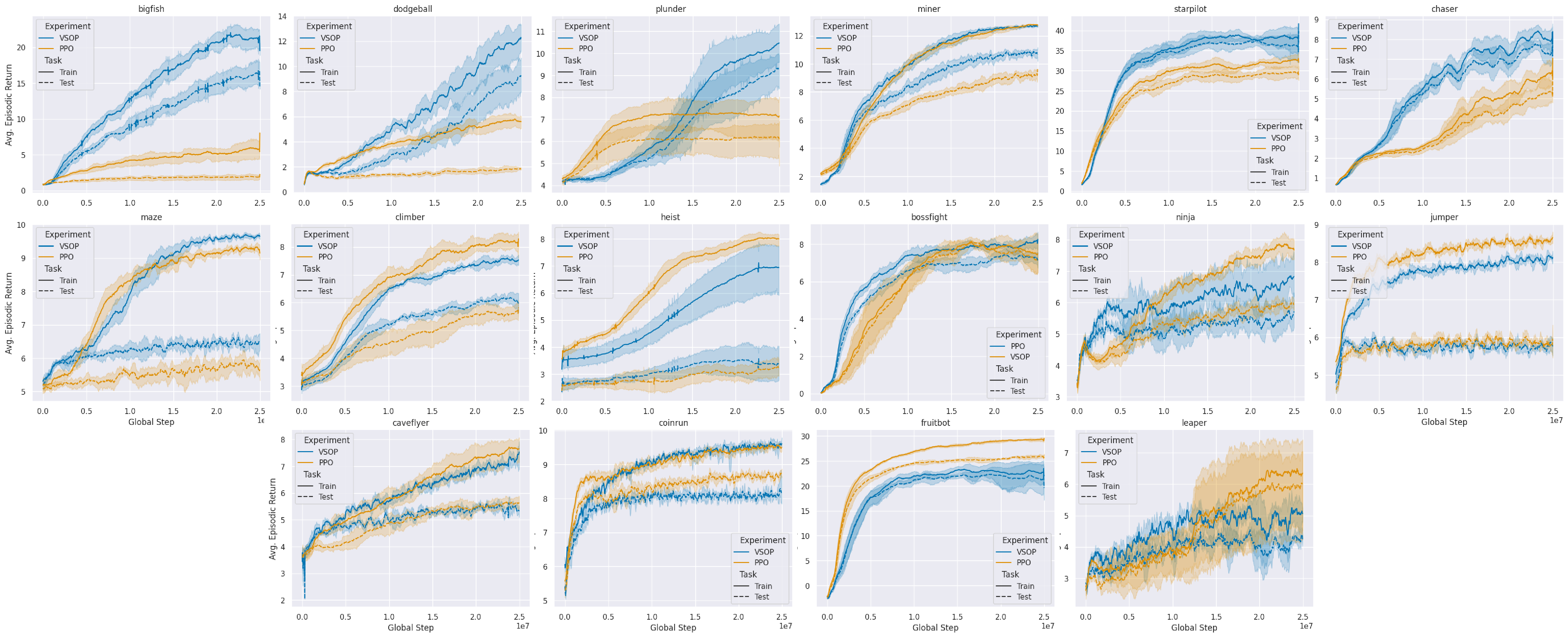}
    \caption{ProcGen training and test curves. We see significant improvement in test set performance on 8 environments, statistical equivalence on 5 environments, and VSOP trails PPO on just 3 environments.}
    \label{fig:procgen-graphs}
\end{figure*}

\subsection{ProcGen}
\label{sec:exp-procgen}
In lieu of finding a suitable benchmark for continuous control, we assess the capacity of VSOP to generalize to unseen environments using ProcGen \citep{cobbe2020leveraging}. ProcGen is a set of 16 environments where game levels are procedurally generated, creating a virtually unlimited set of unique levels. 
We follow the ``easy'' generalization protocol where, for a given environment, models are trained on 200 levels for 25 million time steps and evaluated on the full distribution of environments.
We use the same architecture as PPO in the CleanRL library, and do a Bayesian optimization hyper-parameter search using the bossfight environment.
We search over the learning rate, GAE $\lambda$, number of minibatches per epoch, number of epochs per rollout, the dropout rate, and the entropy regularization coefficient.
Full implementation details are given in \Cref{app:implementation-details-procgen}.
\Cref{fig:rliable-procgen} summarizes our results. 
We see broad significant improvement over PPO across both the PPO and Min-Max normalized metrics.
Furthermore, we see improvement over EDE \citep{jiang2023importance} with respect to the IQM, mean, and optimality gap metrics.
Figure \Cref{fig:procgen-graphs} shows training and test curves for the average episodic return against the number of environment interactions for each environment.
These results present strong evidence for the suitability of VSOP for deployment in non-stationary environments and lend further evidence for the hypothesis of \citet{jiang2023importance} that exploration plays a significant positive role in generalization.

\section{Conclusion}

This work represents a step towards principled approximate Bayesian inference in the on-policy actor-critic setting. Our method is realized through simple modifications to the A3C algorithm, optimizes a lower bound on value plus an additive term and integrates \emph{adaptive state-aware} exploration via Thompson sampling. Our empirical evaluations across several diverse benchmarks confirm our approach's improved performance compared to existing on-policy algorithms.

Establishing convergence rates, especially compared to other algorithms \citep{shen2023towards}, remains an important next step. 
Incorporating theory related to asynchronous settings \citep{shen2023towards} could improve our results, seeing as our most significant efficiency gains are in the single-threaded setting.
Finally, since VSOP does not have significant computational overhead when compared to PPO, it would be interesting to evaluate whether our observed performance gains translate to contemporary large-scale settings where PPO is deployed, such as RLHF \citep{ziegler2019fine}, online gaming \citep{berner2019dota}, or robotics \citep{andrychowicz2020learning}.

\section*{Acknowledgments}
AJ would like to thank Luisa Zintgraf and Panagiotis Tigas for the crash course in reinforcement learning. The authors would like to thank everyone who engaged with this \href{https://twitter.com/anndvision/status/1622915369131180033?s=46&t=MBxzmV7t6dGtGBUvQsCDtg}{Twitter thread}. Specifically, we would like to thank Johan Ferret for highlighting Self-Imitation Advantage Learning, Wilka Carvalho for highlighting  Self-Imitation Learning, Nathan Grinsztajn for highlighting Risk Seeking Policy Gradients, Ohad Rubin for highlighting Discovered Policy Optimization, and Marc Lanctot for the detailed discussion on Regret Matching Policy Gradients. The authors would like to thank Jannik Kossen for brainstorming the title and Nicolas Beltran for contributing to the $C_\pi(\mathbf{s})$ bound. Finally, the authors thank Jacob Beck and all anonymous reviewers for their valuable feedback and suggestions.

\section*{Impact Statement}
Algorithmic decision-making is becoming increasingly present in many areas of our life. 
While this has the potential for benefit, it is also known to automate and perpetuate historical patterns that are often unjust and discriminatory \citep{buolamwini2018gender,noble2018algorithms,benjamin2020race,birhane2021algorithmic}. 
We have proposed an RL algorithm and demonstrated empirically in simulated environments that it can achieve desired capabilities in a manner that requires fewer interactions with the environment when compared with existing algorithms. 
We believe that improving such algorithmic efficiency can help with harm reduction, but we acknowledge that technological solutions alone will not suffice.


\bibliography{references}
\bibliographystyle{icml2024}

\newpage
\appendix
\onecolumn

\section{Supplementary Related Works}
\label{app:related-works}

\begin{table}[ht]
    \caption{Comparison of performance functions for on-policy methods}
    \label{table:performance_metrics_full}
    \centering
    \begin{tabular}{@{}lll@{}}
        \toprule
        Method  & \multicolumn{2}{c}{$\nabla J(\paramp)$} \\ \midrule
        A3C     & $\E_{\pi}\left[ h_{\pi}(\St_{\step}, \At_{\step}) \nabla \log \pi(\At_{\step} \mid \St_{\step}, \paramp) \right];$ & $h_{\pi}(\St_{\step}, \At_{\step}) = q_{\pi}(\St_{\step}, \At_{\step}) - v_{\pi}(\St_{\step})$ \\
        \textbf{VSOP}    & $\E_{\pi}\left[ h_{\pi}^+(\St_{\step}, \At_{\step}) \nabla \log \pi(\At_{\step} \mid \St_{\step}, \paramp) \right];$     & $h_{\pi}^+(\St_{\step}, \At_{\step}) = \max\big(0, h_{\pi}(\St_{\step}, \At_{\step})\big)$ \\
        RMPG    & $\E_{\pi}\left[ \int h_{\pi}^+(\St_{\step}, \at) \nabla d\Pi(\at \mid \St_{\step}, \paramp) \right]$ & \\
        TRPO    & $\E_{\pi}\left[ h_{\pi}(\St_{\step}, \At_{\step}) \nabla \rho(\St_{\step}, \At_{\step}, \paramp) \right];$ & $\rho(\St_{\step}, \At_{\step}, \paramp) = \frac{\pi(\At_{\step} \mid \St_{\step}, \paramp)}{\pi(\At_{\step} \mid \St_{\step}, \paramp_{\mathrm{old}})}$ \\
        PPO     & \multicolumn{2}{l}{$\E_{\pi}\left[ \min \bigg( h_{\pi}(\St_{\step}, \At_{\step}) \nabla \rho(\St_{\step}, \At_{\step}, \paramp), \textrm{clip}\Big( h_{\pi}(\St_{\step}, \At_{\step}) \nabla \rho(\St_{\step}, \At_{\step}, \paramp), 1-\epsilon, 1 + \epsilon \Big) \bigg) \right]$} \\
        DPO & \multicolumn{2}{l}{
            $\E_{\pi}\left[ \nabla \begin{cases}
                \big( h_{\pi}(\rho(\paramp) - 1) - a \tanh( h_{\pi}(\rho(\paramp) - 1)/a ) \big)^+ & h_{\pi}(\St_{\step}, \At_{\step}) \geq 0 \\
                \big( h_{\pi}\log(\rho(\paramp)) - b \tanh(h_{\pi}\log(\rho(\paramp) / b) \big)^+ & h_{\pi}(\St_{\step}, \At_{\step}) < 0 \\
            \end{cases} \right]$
        } \\
        CVaR    & $\E_{\pi}\left[ \big( \nu_{\alpha} - \Gt_{\step}   \big)^+ \nabla \log \pi(\At_{\step} \mid \St_{\step}, \paramp) \right];$ & $\nu_{\alpha} \coloneqq \alpha\text{-quantile of return, } \Gt_{\step}$ \\
        RSPG    & $\E_{\pi}\left[ \big( \Gt_{\step} - \nu_{\alpha} \big)^+ \nabla \log \pi(\At_{\step} \mid \St_{\step}, \paramp) \right];$ &  $\Gt_{\step} \coloneqq \sum_{\kstep = \step + 1}^{\Step} \gamma^{\kstep - 1 - \step} \Rt_{\kstep}$ \\
        EPOpt   & $\E_{\pi}\left[ \mathds{1}\big(\Gt_{\step} \leq \nu_{\alpha}\big) \nabla J(\paramp, \St_{\step}, \At_{\step}) \right];$ & $J(\paramp, \St_{\step}, \At_{\step})$ on-policy perf. function  \\
        \bottomrule
    \end{tabular}
\end{table}
\textbf{Off-policy Methods with Clipped Advantages.}
\emph{Self Imitation Learning (SIL)} \citep{oh2018self} is a hybrid method that uses clipped advantage estimates to improve the performance of on-policy algorithms such as PPO and A2C by learning from its successful off-policy trajectories. By leveraging experience replay, SIL encourages the agent to imitate its high-reward actions. \emph{Self Imitation Advantage Learning (SIAL)} \citep{ferret2020self} extends SIL to the off-policy domain. SIAL uses the clipped advantage function to weigh the importance of different actions during self-imitation, enabling the agent to focus on actions that yield higher long-term rewards. Importantly, even though SIL and SIAL only update policies when advantage estimates are positive, they differ from VSOP in that they are off-policy algorithms that learn from successful past trajectories and optimize different objectives based on max-entropy reinforcement learning \citep{aghasadeghi2011maximum,haarnoja2018soft}.

\textbf{Mirror Learning.}
\emph{Trust Region Policy Optimization (TRPO)} \citep{schulman2015trust} is an on-policy, actor-critic method that improves upon the baseline policy gradient method by incorporating a constraint on the maximum size of policy updates. TRPO takes small steps toward improvement and limits the step size to ensure that the new policy does not deviate significantly from the old policy. TRPO achieves this by optimizing a surrogate objective function that approximates the expected reward under the new policy while imposing a constraint on the KL divergence between the new and old policies. TRPO is effective in various high-dimensional and continuous control tasks. 

\textbf{Risk Sensitive Reinforcement Learning.}
Instead of optimizing expected value, risk-sensitive RL methods optimize a risk measure.
\citet{tamar2015optimizing} propose the risk-averse \emph{CVaR-PG} which seeks to minimize the Conditional Value at Risk (CVaR), $\Phi(\theta) \coloneqq \E_{\pi} \left[ \Gt_{\step} \mid \Gt_{\step} \leq \nu_{\alpha} \right]$, where $\nu_{\alpha}$ is the $\alpha$-quantile of the return, $\Gt_{\step}$, distribution under the policy, $\pi(\at \mid \st,  \paramp)$.
Relatedly, \citet{tang2020worst} have used the CVaR as a baseline function for standard policy updates.
By focusing only on the worse case trajectories, CVaR-PG is susceptible to ``blindness to success,'' thus \citet{greenberg2022efficient} propose a Cross-entropy Soft-Risk algorithm (CeSoR) to address this.
\citet{kenton2019generalizing} and \citet{filos2022model} also propose uncertainty aware, risk-averse methods.
For model-based policy gradient methods, \citet{rajeswaran2016epopt} propose \emph{Ensemble Policy Optimization (EPOpt)}, which incorporates restricting policy updates to be risk-averse based on the CVaR and uses ensembles to sample hypothesized models.
In contrast to the above risk-averse methods, \citet{petersen2019deep} present \emph{Risk Seeking Policy Gradient (RSPG)} which focuses on maximizing best-case performance by only performing gradient updates when rewards exceed a specified quantile of the reward distribution.
\citet{prashanth2022risk} provide a comprehensive discussion on risk-sensitive RL.

\section{Theoretical Results}

\subsection{Proof of \Cref{th:main}}
\label{app:constant-proof}

\begin{theorem}
    Let, $\Gt_{\step} \coloneqq \sum_{\kstep = \step + 1}^{\Step} \gamma^{\kstep - 1 - \step}  \Rt_{\kstep}$, denote the discounted return.
    Let $q_{\pi}(\st, \at) = \E_{\pi} \left[ \Gt_{\step} \mid \St_{\step} = \st,  \At_{\step} = \at\right]$, denote the state-action value function, and
    $v_{\pi}(\st) = \E_{\pi} \left[ \Gt_{\step} \mid \St_{\step} = \st \right]$, denote the state value function, under policy $\pi(\at \mid \st, \paramp)$.
    Define the ReLU function as $\big(x\big)^+ \coloneqq \max(0, x)$.
    Assume that rewards, $\Rt_{\step}$, are non-negative and the gradient of the policy, $\nabla \pi(\at \mid \st, \paramp)$, is a conservative vector field.
    Then, performing gradient ascent with respect to,
    \begin{equation}
        \nabla_{\paramp} J(\paramp) = \E_{\pi}\left[ \Bigl(q_{\pi}(\St_{\step}, \At_{\step}) - v_{\pi}(\St_{\step}) \Bigr)^+ \nabla_{\paramp} \log \pi(\At_{\step} \mid \St_{\step}, \paramp) \right],
        \label{eq:app-objective}
    \end{equation}
    maximizes a lower-bound, $v_{\pi}^*(\st)$, on the state value function, $v_{\pi}(\st)$, plus an additive term:
    \begin{equation}
        v_{\pi}^*(\st) \leq v_{\pi}(\st) + C_{\pi}(\st),
        \label{eq:app-bound}
    \end{equation}
    where, $C_{\pi}(\st) = \iint \Big( \gamma v_{\pi}(\st') - v_{\pi}(\st) \Big)^{+} d\Prob(\st' \mid \St_{\step}=\st, \At_{\step} = \at) d\Pi(\at \mid \St_{\step}=\st)$, is the expected, clipped difference in the state value function, $\gamma v_{\pi}(\st') - v_{\pi}(\st)$, over all actions, $\at$, and next states, $\st'$, under the policy given state, $\st$.
    Here, we use $\int \dots d\Pi(\at \mid \st)$ to denote $\sum_{\at} \dots \pi(\at \mid \st)$ for discrete action spaces and $\int \dots \pi(\at \mid \st)d\at$ for continuous action spaces.
    Similarly, we use $\int \dots d\Prob(\st' \mid \st, \at)$ to denote $\sum_{\st'} \dots p(\st' \mid \st, \at)$ for discrete state spaces and $\int \dots p(\st' \mid \st, \at)d\st'$ for continuous state spaces.

    \begin{proof}
        \Cref{cor:pg_decomposition} shows that the policy-gradient theorem \citep{sutton1999policy} can be expressed in terms of the clipped advantage function,
        \[
            \mathcolor{mypurple}{h_{\pi}^{+}(\st, \at)} = \mathcolor{mypurple}{\Bigl(q_{\pi}(\st, \at) - v_{\pi}(\st) \Bigr)^+} \coloneqq \mathcolor{mypurple}{\max(0, q_{\pi}(\st, \at) - v_{\pi}(\st))},
        \]
        as,
        \begin{equation}
            \begin{split}
                \nabla v_{\pi}(\st)  
                &= \int_{\Scal} \sum_{k=0}^{\infty} \Bigg[ \gamma^k \int_{\Acal} \mathcolor{mypurple}{h_{\pi}^{+}(\mathbf{x}, \at)} \nabla d\Pi(\at \mid \mathbf{x}) \Bigg] d\prob(\st \to \mathbf{x}; k, \pi) \\
                &\quad\quad + \int_{\Scal} \sum_{k=0}^{\infty} \Bigg[ \gamma^k \int_{\Acal} \mathds{1}\big(q_{\pi}(\mathbf{x}, \at) > v_{\pi}(\mathbf{x}) \big) v_{\pi}(\mathbf{x}) \nabla d\Pi(\at \mid \mathbf{x}) \Bigg] d\prob(\st \to \mathbf{x}; k, \pi) \\
                &\quad\quad\quad\quad + \int_{\Scal} \sum_{k=0}^{\infty} \Bigg[ \gamma^k \int_{\Acal} \mathds{1}\big(q_{\pi}(\mathbf{x}, \at) \leq v_{\pi}(\mathbf{x}) \big) q_{\pi}(\mathbf{x}, \at) \nabla d\Pi(\at \mid \mathbf{x}) \Bigg] d\prob(\st \to \mathbf{x}; k, \pi),
            \end{split}
        \end{equation}
        where, $\Prob(\st \to \mathbf{x}; k, \pi)$, is the probability of transitioning from state $\st$ to state $\mathbf{x}$ in $k$ steps under policy $\pi$. 
        
        The first right hand side term above defines the gradient of the lower-bound, $v_{\pi}^*(\st)$, with respect to $\paramp$:
        \begin{equation}
            \nabla v_{\pi}^*(\st) \coloneqq \int_{\Scal} \sum_{k=0}^{\infty} \Bigg[ \gamma^k \int_{\Acal} \mathcolor{mypurple}{h_{\pi}^{+}(\mathbf{x}, \at)} \nabla d\Pi(\at \mid \mathbf{x}) \Bigg] d\prob(\st \to \mathbf{x}; k, \pi).
        \end{equation}
        
        Letting, $\nabla v_{\pi}^{*}(\st_0)=\int_{\Scal} \sum_{k=0}^{\infty}  \gamma^k \int_{\Acal} h_{\pi}^{+}(\st, \at) \nabla d\Pi(\at \mid \st) d\Prob(\st_0 \to \st; k, \pi)$, a straightforward continuation of the policy gradient theorem \citep{sutton1999policy} will show that 
        \[
            \nabla J(\paramp) \coloneqq \nabla v_{\pi}^{*}(\st_0) \propto \iint h_{\pi}^{+}(\st, \at) \nabla_{\paramp} d\Pi(\at \mid \st, \paramp) d\mathrm{P}(\st).
        \] 

        We then arrive at \Cref{eq:app-objective} by moving from the all states/actions to single state/action formulation:
        \begin{subequations}
            \begin{align*}
                \nabla J(\paramp) 
                &\coloneqq \nabla v_{\pi}^{*}(\st_0),
                && \text{by definition} \\
                &\propto \iint \Bigl(q_{\pi}(\st, \at) - v_{\pi}(\st) \Bigr)^+ \nabla_{\paramp} d\Pi(\at \mid \st, \paramp) d\mathrm{P}(\st),
                && \text{\citet{sutton1999policy}} \\
                &= \E_{\pi} \left[ \int \Bigl(q_{\pi}(\St_{\step}, \at) - v_{\pi}(\St_{\step}) \Bigr)^+ \nabla_{\paramp} d\Pi(\at \mid \St_{\step}, \paramp) \right],
                && \text{} \\
                &= \E_{\pi} \left[ \int \Bigl(q_{\pi}(\St_{\step}, \at) - v_{\pi}(\St_{\step}) \Bigr)^+ \frac{\nabla_{\paramp} d\Pi(\at \mid \St_{\step}, \paramp)}{d\Pi(\at \mid \St_{\step}, \paramp} d\Pi(\at \mid \St_{\step}, \paramp \right],
                && \text{} \\
                &= \E_{\pi} \left[ \int \Bigl(q_{\pi}(\St_{\step}, \At_{\step}) - v_{\pi}(\St_{\step}) \Bigr)^+ \nabla_{\paramp} \log{\pi(\At_{\step} \mid \St_{\step}, \paramp)} \right].
                && \text{}
            \end{align*}
        \end{subequations}
        
        Now we need to show that,
        \[
            v_{\pi}^*(\st) \leq v_{\pi}(\st) + \iint \Big( \gamma v_{\pi}(\st') - v_{\pi}(\st) \Big)^{+} d\Prob(\st' \mid \St_{\step}=\st, \At_{\step}) d\Pi(\at \mid \St_{\step}=\st).
        \]
        To do so, we will first prove that it holds for episodes, $\Step$, of length 1, then that it holds for episodes of length 2.
        These two proofs will then prove \Cref{eq:app-bound} for episodes of arbitrary length by mathematical induction and conclude the proof.

        \textbf{For episodes of length 1, $|T| = 1$}, we have
        \begin{equation}
            \begin{split}
                \nabla v_{\pi}(\st) 
                &= \int q_{\pi}(\st, \at) \nabla d\Pi(\at \mid \st) + \int \nabla q_{\pi}(\st, \at) d\Pi(\at \mid \st), \\
                &= \int q_{\pi}(\st, \at) \nabla d\Pi(\at \mid \st) + \int \bigg( \nabla \int \rt d\Prob(\rt \mid \st, \at) \bigg) d\Pi(\at \mid \st), \\
                &= \int q_{\pi}(\st, \at) \nabla d\Pi(\at \mid \st), \\
                &= \int h_{\pi}^{+}(\st, \at) \nabla d\Pi(\at \mid \st) + \int \Big( \mathds{1}\big(q_{\pi} > v_{\pi} \big) v_{\pi}(\st) + \mathds{1}\big(q_{\pi} \leq v_{\pi} \big) q_{\pi}(\st, \at) \Big) \nabla d\Pi(\at \mid \st).
            \end{split}
        \end{equation}
        Therefore, for $|T| = 1$,
        \[
            \nabla v_{\pi}^*(\st) = \int h_{\pi}^{+}(\st, \at) \nabla d\Pi(\at \mid \st)
        \]

        In order to recover $v_{\pi}^*(\st)$, we need to use the work of \citet{willse2019inverse} to define an inverse function for the gradient.
        Assume that the policy, $\pi(\at \mid \st, \paramp)$, is a smooth, infinitely differentiable function with respect to $\paramp$. Further, let the gradient of the policy,
        \begin{equation}
            \nabla \pi(\at \mid \st, \paramp) =
            \begin{pmatrix}
                \frac{\partial}{\partial\theta_1} \pi(\at \mid \st, \theta_1), \\
                \vdots \\
                \frac{\partial}{\partial\theta_k} \pi(\at \mid \st, \theta_k)
            \end{pmatrix},
        \end{equation}
        be a conservative vector field. We call $\tilde{\beta}\big(\nabla \pi(\at \mid \st, \paramp)\big)$ the inverse of the gradient operation, $\nabla \pi(\at \mid \st, \paramp)$.
        Assuming that $\pi(\at \mid \st, \paramp)$ is a representative of $\tilde{\beta}$, we have that,
        \begin{equation}
            \label{eq:grad-inverse}
            \begin{split}
                \pi(\at \mid \st, \paramp) &= \tilde{\beta}\big(\nabla \pi(\at \mid \st, \paramp)\big), \\
                &= \int_{\gamma} \nabla \pi(\at \mid \st, \paramp) d\mathbf{x}, \\
                &= \int_{\gamma} \frac{\partial}{\partial\theta_1} \pi(\at \mid \st, \theta_1) d\theta_1 + \dots + \frac{\partial}{\partial\theta_k} \pi(\at \mid \st, \theta_k) d\theta_k,
            \end{split}
        \end{equation}
        where $\gamma$ is a path from the fixed reference point, $\paramp_0$, to $\paramp$. The conservativeness of $\nabla \pi(\at \mid \st, \paramp)$ guarantees that the integrals are path independent.

        Now we have,
        \begin{subequations}
            \begin{align*}
                v_{\pi}^*(\st) 
                &= \tilde{\beta}\bigg( \int h_{\pi}^{+}(\st, \at) \nabla d\Pi(\at \mid \st) \bigg), \\
                &=  \int h_{\pi}^{+}(\st, \at) \tilde{\beta}\big(\nabla d\Pi(\at \mid \st) \big),
                && \text{linearity} \\
                &=  \int h_{\pi}^{+}(\st, \at) d\Pi(\at \mid \st), 
                && \text{\Cref{eq:grad-inverse}}\\
                &\leq \iint \Big( \rt + \big( \gamma v_{\pi}(\st') - v_{\pi}(\st) \big)^+ \Big) d\Prob(\st', \rt \mid \st, \at)d\Pi(\at \mid \st),
                && \text{\Cref{cor:lower-bound-decomp}} \\
                &= v_{\pi}(\st) + \iint \big( \gamma v_{\pi}(\st') - v_{\pi}(\st) \big)^+ d\Prob(\st' \mid \st, \at)d\Pi(\at \mid \st),
                && \text{$|T| = 1$}
            \end{align*}
        \end{subequations}
        which concludes the proof for episodes of length 1.

        \textbf{For episodes of length 2}, $|T| = 2$, we have
        \begin{subequations}
            \begin{align*}
                \nabla v_{\pi}(\st) 
                &= \int q_{\pi}(\st, \at) \nabla d\Pi(\at \mid \st) + \int \nabla q_{\pi}(\st, \at) d\Pi(\at \mid \st), \\
                \begin{split}
                    &= \int q_{\pi}(\st, \at) \nabla d\Pi(\at \mid \st) + \iiint q_{\pi}(\st', \at') \nabla d\Pi(\at' \mid \st') d\Prob(\st' \mid \at, \st) d\Pi(\at \mid \st) \\
                    &\quad\quad+ \iiint \bigg( \nabla \int \rt' d\Prob(\rt' \mid \st', \at') \bigg) d\Pi(\at' \mid \st'),
                \end{split} \\
                &= \int q_{\pi}(\st, \at) \nabla d\Pi(\at \mid \st) + \iiint q_{\pi}(\st', \at') \nabla d\Pi(\at' \mid \st') d\Prob(\st' \mid \at, \st) d\Pi(\at \mid \st), \\
                \begin{split}
                    &= \int h_{\pi}^{+}(\st, \at) \nabla d\Pi(\at \mid \st) + \iiint h_{\pi}^{+}(\st', \at') \nabla d\Pi(\at' \mid \st') d\Prob(\st' \mid \at, \st) d\Pi(\at \mid \st) \\
                    &\quad\quad+ \int \Big( \mathds{1}\big(q_{\pi} > v_{\pi} \big) v_{\pi}(\st) + \mathds{1}\big(q_{\pi} \leq v_{\pi} \big) q_{\pi}(\st, \at) \Big) \nabla d\Pi(\at \mid \st) \\
                    &\quad\quad+ \iiint \Big( \mathds{1}\big(q_{\pi} > v_{\pi} \big) v_{\pi}(\st') + \mathds{1}\big(q_{\pi} \leq v_{\pi} \big) q_{\pi}(\st', \at') \Big) \nabla d\Pi(\at' \mid \st')d\Prob(\st' \mid \at, \st) d\Pi(\at \mid \st).
                \end{split}
            \end{align*}
        \end{subequations}
        Therefore, for $|T| = 2$,
        \[
            \nabla v_{\pi}^*(\st) = \int h_{\pi}^{+}(\st, \at) \nabla d\Pi(\at \mid \st) + \iiint h_{\pi}^{+}(\st', \at') \nabla d\Pi(\at' \mid \st') d\Prob(\st' \mid \at, \st) d\Pi(\at \mid \st).
        \]
        Finally, we apply the $\tilde{\beta}$ operator:
        \begin{subequations}
            \begin{align*}
                v_{\pi}^*(\st) 
                &= \tilde{\beta}\bigg(\int h_{\pi}^{+}(\st, \at) \nabla d\Pi(\at \mid \st) + \iiint h_{\pi}^{+}(\st', \at') \nabla d\Pi(\at' \mid \st') d\Prob(\st' \mid \at, \st) d\Pi(\at \mid \st)\bigg), \\
                &= \int h_{\pi}^{+}(\st, \at) \tilde{\beta}\Big(\nabla d\Pi(\at \mid \st)\Big) + \iiint h_{\pi}^{+}(\st', \at') \tilde{\beta}\Big(\nabla d\Pi(\at' \mid \st')\Big) d\Prob(\st' \mid \at, \st) d\Pi(\at \mid \st), 
                && \text{linearity} \\
                &= \int h_{\pi}^{+}(\st, \at) d\Pi(\at \mid \st) + \iiint h_{\pi}^{+}(\st', \at') d\Pi(\at' \mid \st') d\Prob(\st' \mid \at, \st) d\Pi(\at \mid \st), 
                && \text{\Cref{eq:grad-inverse}} \\
                \begin{split}
                    &\leq \iint \rt d\Prob(\rt \mid \st, \at) d\Pi(\at \mid \st) + \iint \big( \gamma v_{\pi}(\st') - v_{\pi}(\st) \big)^+ d\Prob(\st' \mid \st, \at)d\Pi(\at \mid \st) \\
                    &\quad\quad + \iiint h_{\pi}^{+}(\st', \at') d\Pi(\at' \mid \st') d\Prob(\st' \mid \at, \st) d\Pi(\at \mid \st),
                \end{split}
                && \text{\Cref{cor:lower-bound-decomp}} \\
                \begin{split}
                    &\leq \iint \rt d\Prob(\rt \mid \st, \at) d\Pi(\at \mid \st) + \iint \big( \gamma v_{\pi}(\st') - v_{\pi}(\st) \big)^+ d\Prob(\st' \mid \st, \at)d\Pi(\at \mid \st) \\
                    &\quad\quad + \iint \gamma v_{\pi}(\st') d\Prob(\st' \mid \at, \st) d\Pi(\at \mid \st),
                \end{split}
                && \text{\Cref{cor:lower-bound}} \\
                &= v_{\pi}(\st) + \iint \big( \gamma v_{\pi}(\st') - v_{\pi}(\st) \big)^+ d\Prob(\st' \mid \st, \at)d\Pi(\at \mid \st).
                && \text{rearranging terms}
            \end{align*}
        \end{subequations}
        
    \end{proof}
\end{theorem}

\textbf{Negative Rewards Add a Known Constant to the Lower Bound, but Do Not Impact Optimization.}
Consider the scenario where rewards can be negative ($r_{\text{min}} \leq 0$), and we adjust observed rewards by subtracting $r_{\text{min}}$. This adjustment leads to the following expressions for the state-value and action-value functions, respectively:

$$ v_{\pi}^{r_{\text{min}}}(s) := E_\pi\Bigg[\sum^T_{k=t+1}\gamma^{k-t-1}(R_k - r_{\text{min}}) \mid S_t=s \Bigg] = v_{\pi}(s) - \sum^T_{k=t+1}\gamma^{k-t-1}r_{\text{min}} $$

$$ q_{\pi}^{r_{\text{min}}}(s, a) := E_\pi\Bigg[\sum^T_{k=t+1}\gamma^{k-t-1}(R_k - r_{\text{min}}) \mid S_t=s, A_t=a \Bigg] = q_{\pi}(s, a) - \sum^T_{k=t+1}\gamma^{k-t-1}r_{\text{min}} $$ In both episodic ($0< T < \infty$) and continuing ($T \to \infty$) cases, the transformations yield:

$$ v_{\pi}^{r_{\text{min}}} (s) \leq v_{\pi}(s) - \frac{\gamma}{1-\gamma}r_{\text{min}}, \quad q_{\pi}^{r_{\text{min}}}(s, a) \leq q_{\pi}(s, a) - \frac{\gamma}{1-\gamma}r_{\text{min}}. $$

Crucially, the advantage function remains invariant under this transformation:
\begin{equation}
    \begin{split}
        h_{\pi}^{r_{\text{min}}} (s, a) &= q_{\pi}^{r_{\text{min}}}(s, a) - v^{r_{\text{min}}}_{\pi}(s) \\
        &= q_{\pi}(s, a) - \sum^T_{k=t+1}\gamma^{k-t-1}r_{\text{min}} - \Bigg(v_{\pi}(s) - \sum^T_{k=t+1}\gamma^{k-t-1}r_{\text{min}}\Bigg) \\
        &= q_{\pi}(s, a) - v_{\pi}(s) \\
        &= h_{\pi}(s, a)
    \end{split}
\end{equation}

Imagine now that we use the transformed advantage function $h^{r_{\text{min}}}_{\pi}(s, a)$ to ensure non-negativity in our proof for Theorem 3.1. Then, our objective maximizes:
\begin{equation}
    \begin{split}
        v_\pi^{*,r_{\text{min}}} (s) &= v_{\pi}^{r_{\text{min}}}(s) + C^{r_{\text{min}}}_\pi(s) \\
        &= v_{\pi}^{r_{\text{min}}} (s) + \iint \big(\gamma v_{\pi}^{r_{\text{min}}}(s') - v^{r_{\text{min}}}_{\pi}(s)\big)^+dP(s' \mid s, a)d\Pi(a \mid s)
    \end{split}
\end{equation}
Let's focus on the integrand of the $C^{r_{\text{min}}}_\pi(s)$ term:

\begin{equation}
    \begin{split}
        \Big( \gamma v_\pi^{r_\text{min}} (s') - v^{r_\text{min}}_\pi (s) \Big)^+ &= \bigg(\gamma v_{\pi}(s') - v_{\pi}(s) + \sum_{k=t+1}^T\gamma^{k-1-t}r_{\text{min}} - \gamma \sum_{k=t+2}^T\gamma^{k-1-t}r_{\text{min}}\bigg)^+ \\
        &= \Big(\gamma v_{\pi}(s') - v_{\pi}(s) + r_{\text{min}} + \gamma r_{\text{min}} - \gamma^{T-t} r_{\text{min}}\Big)^+ \\
        &= \Big(\gamma v_{\pi}(s') - v_{\pi}(s) + c_t\Big)^+: c_t < 0 \\
        &= \frac{1}{2} \Big( \gamma v_{\pi}(s') - v_{\pi}(s) + c_t + |\gamma v_{\pi}(s') - v_{\pi}(s) + c_t| \Big) \\
        &\leq \Big(\gamma v_{\pi}(s') - v_{\pi}(s)\Big)^+ + \frac{1}{2}(c_t + |c_t|) \\
        &= \Big(\gamma v_{\pi}(s') - v_{\pi}(s)\Big)^+
    \end{split}
\end{equation}

Therefore, $C_\pi^{r_{\text{min}}} (s) <= C_\pi(s)$, which maintains the direction of the lower bound. So, should the rewards be negative, we optimize
$$v_\pi^{*,r_{\text{min}}} (s) \leq v_\pi(s) - \frac{\gamma}{1-\gamma}r_{\text{min}} + C_\pi(s).$$

This expression has an extra non-negative (yet fully determined) constant $-\frac{\gamma}{1-\gamma}r_{\text{min}}$.
Importantly, this result has no bearing on the optimization procedure. 
The invariance of the advantage function ensures that the performance function is also invariant to a shift in rewards: $$ \nabla_\theta J(\theta) := E_{\pi}\bigg[ \Bigl(q_{\pi}(S_{t}, A_{t}) - v_{\pi}(S_{t}) \Bigr)^+ \nabla_{\theta} \log \pi(A_{t} \mid S_{t}, \theta) \bigg] = E_{\pi}\bigg[ \Bigl(q_{\pi}^{r_{\text{min}}}(S_{t}, A_{t}) - v_{\pi}^{r_{\text{min}}}(S_{t}) \Bigr)^+ \nabla_{\theta} \log \pi(A_{t} \mid S_{t}, \theta) \bigg], $$ which entails that no special measures need to be taken when rewards can take negative values.

\begin{lemma}
    \label{cor:pg_decomposition}
    $\nabla v_{\pi}(\st)$ can be written in terms of $h_{\pi}^{+}(\st, \at)$.
    \begin{proof}
        \begin{subequations}
            \begin{align}
                \nabla v_{\pi}(\st) 
                &= \nabla \bigg[ \int q_{\pi}(\st, \at) d\Pi(\at \mid \st) \bigg], \label{eq:three_goats-a} \\
                &= \int q_{\pi}(\st, \at) \nabla d\Pi(\at \mid \st) + \int \nabla q_{\pi}(\st, \at) d\Pi(\at \mid \st), \label{eq:three_goats-b} \\
                \begin{split}
                    &= \int \Big( \mathcolor{mygreen}{h_{\pi}^{+}(\st, \at)} + \mathcolor{myfuchsia}{\mathds{1}\big(q_{\pi} > v_{\pi} \big) v_{\pi}(\st)} + \mathcolor{mypurple}{\mathds{1}\big(q_{\pi} \leq v_{\pi} \big) q_{\pi}(\st, \at)} \Big) \nabla d\Pi(\at \mid \st) \\
                    &\quad\quad + \int \nabla q_{\pi}(\st, \at) d\Pi(\at \mid \st),
                \end{split} \label{eq:three_goats-c} \\
                \begin{split}
                    &= \int \Big( \mathcolor{mygreen}{h_{\pi}^{+}(\st, \at)} + \mathcolor{myfuchsia}{\mathds{1}\big(q_{\pi} > v_{\pi} \big) v_{\pi}(\st)} + \mathcolor{mypurple}{\mathds{1}\big(q_{\pi} \leq v_{\pi} \big) q_{\pi}(\st, \at)} \Big) \nabla d\Pi(\at \mid \st)  \\
                    &\quad\quad + \int  \nabla \bigg[ \int  \big( \rt + \gamma v_{\pi}(\st') \big) d\prob(\st', \rt \mid \st, \at) \bigg] d\Pi(\at \mid \st),
                \end{split} \label{eq:three_goats-d} \\
                \begin{split}
                    &= \int \Big( \mathcolor{mygreen}{h_{\pi}^{+}(\st, \at)} + \mathcolor{myfuchsia}{\mathds{1}\big(q_{\pi} > v_{\pi} \big) v_{\pi}(\st)} + \mathcolor{mypurple}{\mathds{1}\big(q_{\pi} \leq v_{\pi} \big) q_{\pi}(\st, \at)} \Big) \nabla d\Pi(\at \mid \st)  \\
                    &\quad\quad + \gamma\iint \nabla v_{\pi}(\st') d\prob(\st' \mid \st, \at) d\Pi(\at \mid \st),
                \end{split} \label{eq:three_goats-e} \\
                \begin{split}
                    &= \int \Big( \mathcolor{mygreen}{h_{\pi}^{+}(\st, \at)} + \mathcolor{myfuchsia}{\mathds{1}\big(q_{\pi} > v_{\pi} \big) v_{\pi}(\st)} + \mathcolor{mypurple}{\mathds{1}\big(q_{\pi} \leq v_{\pi} \big) q_{\pi}(\st, \at)} \Big) \nabla d\Pi(\at \mid \st) \\
                    &\quad\quad + \gamma \iint \Bigg[ \int q_{\pi}(\st', \at') \nabla d\Pi(\at' \mid \st') \\
                    &\quad\quad\quad\quad + \gamma \int  \nabla v_{\pi}(\st'') d\prob(\st'' \mid \st', \at') d\Pi(\at' \mid \st') \Bigg] d\prob(\st' \mid \st, \at) d\Pi(\at \mid \st),
                \end{split} \label{eq:three_goats-f} \\
                \begin{split}
                    &= \int \Big( \mathcolor{mygreen}{h_{\pi}^{+}(\st, \at)} + \mathcolor{myfuchsia}{\mathds{1}\big(q_{\pi} > v_{\pi} \big) v_{\pi}(\st)} + \mathcolor{mypurple}{\mathds{1}\big(q_{\pi} \leq v_{\pi} \big) q_{\pi}(\st, \at)} \Big) \nabla d\Pi(\at \mid \st) \\
                    &\quad\quad + \gamma \iint \Bigg[ \int  \Big( \mathcolor{mygreen}{h_{\pi}^{+}(\st', \at')} + \mathcolor{myfuchsia}{\mathds{1}\big(q_{\pi} > v_{\pi} \big) v_{\pi}(\st')} + \mathcolor{mypurple}{\mathds{1}\big(q_{\pi} \leq v_{\pi} \big) q_{\pi}(\st', \at')} \Big) \nabla d\Pi(\at' \mid \st') \\
                    &\quad\quad\quad\quad + \gamma \int \nabla v_{\pi}(\st'') d\prob(\st'' \mid \st', \at') d\Pi(\at' \mid \st') \Bigg] d\prob(\st' \mid \st, \at) d\Pi(\at \mid \st),
                \end{split} \label{eq:three_goats-g} \\
                \begin{split}
                    &= \int_{\Scal} \sum_{k=0}^{\infty} \Bigg[ \gamma^k \int_{\Acal} \mathcolor{mygreen}{h_{\pi}^{+}(\mathbf{x}, \at)} \nabla d\Pi(\at \mid \mathbf{x}) \Bigg] d\prob(\st \to \mathbf{x}; k, \pi) \\
                    &\quad\quad + \int_{\Scal} \sum_{k=0}^{\infty} \Bigg[ \gamma^k \int_{\Acal} \mathcolor{myfuchsia}{\mathds{1}\big(q_{\pi}(\mathbf{x}, \at) > v_{\pi}(\mathbf{x}) \big) v_{\pi}(\mathbf{x})} \nabla d\Pi(\at \mid \mathbf{x}) \Bigg] d\prob(\st \to \mathbf{x}; k, \pi) \\
                    &\quad\quad\quad\quad + \int_{\Scal} \sum_{k=0}^{\infty} \Bigg[ \gamma^k \int_{\Acal} \mathcolor{mypurple}{\mathds{1}\big(q_{\pi}(\mathbf{x}, \at) \leq v_{\pi}(\mathbf{x}) \big) q_{\pi}(\mathbf{x}, \at)} \nabla d\Pi(\at \mid \mathbf{x}) \Bigg] d\prob(\st \to \mathbf{x}; k, \pi)
                \end{split} \label{eq:three_goats-h}
            \end{align}
        \end{subequations}
    \end{proof}
\end{lemma}

\begin{lemma}
    \label{cor:lower-bound-decomp}
    \begin{equation*}
        \underline{v}_{\pi}^{v_{\pi}}(\st) \leq \iint \rt d\Prob(\rt \mid \st, \at)d\Pi(\at \mid \st) + \iint \Big( \gamma v_{\pi}(\st') - v_{\pi}(\st) \Big)^+ d\Prob(\st' \mid \st, \at)d\Pi(\at \mid \st)
    \end{equation*}
    \begin{proof}
        \begin{subequations}
            \begin{align*}
                \underline{v}_{\pi}^{v_{\pi}}(\st)
                &\coloneqq \int h_{\pi}^{+}(\st, \at) d\Pi(\at \mid \st) \\
                &= \frac{1}{2} \int \Big( q_{\pi}(\st, \at) - v_{\pi} + \big|q_{\pi}(\st, \at) - v_{\pi} \big| \Big) d\Pi(\at \mid \st)
                && \text{(}2\max(0, a) = a + |a|\text{)} \\
                \begin{split}
                    &= \frac{1}{2} \int \bigg( \int \big( \rt + \gamma v_{\pi}(\st') - v_{\pi}(\st) \big) d\Prob(\st', \rt \mid \st, \at) \\
                    &\quad\quad+ \Big| \int \big( \rt + \gamma v_{\pi}(\st') - v_{\pi}(\st) \big) d\Prob(\st', \rt \mid \st, \at) \Big| \bigg) d\Pi(\at \mid \st)
                \end{split} \\
                \begin{split}
                    &\leq \frac{1}{2}\iint \Big( \rt + \gamma v_{\pi}(\st') - v_{\pi}(\st) + \big| \rt + \gamma v_{\pi}(\st') - v_{\pi}(\st) \big| \Big)  \\
                    &\quad\quad d\Prob(\st', \rt \mid \st, \at) d\Pi(\at \mid \st)
                \end{split} 
                && \text{(Jensen's inequality)} \\
                \begin{split}
                    &\leq \frac{1}{2}\iint \Big( 2\rt + \gamma v_{\pi}(\st') - v_{\pi}(\st) + \big| \gamma v_{\pi}(\st') - v_{\pi}(\st) \big| \Big) \\
                    &\quad\quad d\Prob(\st', \rt \mid \st, \at) d\Pi(\at \mid \st)
                \end{split} 
                && \text{(triangle inequality)} \\
                &= \iint \Big( \rt + \big( \gamma v_{\pi}(\st') - v_{\pi}(\st) \big)^+ \Big) d\Prob(\st', \rt \mid \st, \at)d\Pi(\at \mid \st)
                && \text{(}2\max(0, a) = a + |a|\text{)}
            \end{align*}
        \end{subequations}
    \end{proof}
\end{lemma}

\begin{lemma}
    \label{cor:lower-bound}
    When, without loss of generality, rewards, $\Rt_{\step}$, are assumed to be non-negative:
    \begin{equation*}
        \underline{v}_{\pi}^{v_{\pi}}(\st) \coloneqq \int h_{\pi}^{+}(\st, \at) d\Pi(\at \mid \st) \leq v_{\pi}(\st)
    \end{equation*}
    \begin{proof}
        \begin{subequations}
            \begin{align*}
                \int h_{\pi}^{+}(\st, \at) d\Pi(\at \mid \st)
                &= \frac{1}{2} \int \Big( q_{\pi}(\st, \at) - v_{\pi} + \big|q_{\pi}(\st, \at) - v_{\pi} \big| \Big) d\Pi(\at \mid \st)
                && \text{( } 2\max(0, a) = a + |a| \text{ )} \\
                &\leq \int q_{\pi}(\st, \at) d\Pi(\at \mid \st)
                && \text{(triangle inequality)} \\
                &= v_{\pi}(\st)
            \end{align*}
        \end{subequations}
    \end{proof}
\end{lemma}

\subsection{Relation to Regret Matching Policy Gradient (RMPG)}
\label{sec:rmpg-to-vsop}
Here we provide a derivation starting from RMPG and arriving at our method.

\begin{equation*}
    \begin{split}
        \nabla J(\paramp) 
        &= \mathbb{E}_{\pi}^{} \left[ \int_{\Acal}  \left( q_{\pi}^{}(\St_{\step}, \at) - \int_{\Acal} \pi(\at^{\prime} \mid \St_{\step}, \paramp) q_{\pi}^{}(\St_{\step}, \at^{\prime}) d\at^{\prime} \right)^{+} \nabla_{\paramp} \pi(\at \mid \St_{\step}, \paramp) d\at \right] \\
        &= \mathbb{E}_{\pi}^{} \left[ \int_{\Acal} \left( q_{\pi}^{}(\St_{\step}, \at) - v_{\pi}^{}(\St_{\step}) \right)^{+} \nabla_{\paramp} \pi(\at \mid \St_{\step}, \paramp) d\at \right] \\
        &= \mathbb{E}_{\pi}^{} \left[ \int_{\Acal} h_{\pi}^{+}(\St_{\step}, \at) \nabla_{\paramp} \pi(\at \mid \St_{\step}, \paramp) d\at \right] \\
        &= \mathbb{E}_{\pi}^{} \left[ \int_{\Acal} \pi(\at \mid \St_{\step}, \paramp) h_{\pi}^{+}(\St_{\step}, \at) \frac{\nabla_{\paramp} \pi(\at \mid \St_{\step}, \paramp)}{\pi(\at \mid \St_{\step}, \paramp)} d\at \right] \\
        &= \mathbb{E}_{\pi}^{} \left[ h_{\pi}^{+}(\St_{\step}, \At_{\step}) \frac{\nabla_{\paramp} \pi(\At_{\step} \mid \St_{\step}, \paramp)}{\pi(\At_{\step} \mid \St_{\step}, \paramp)} \right] \\
        &= \E_{\pi}^{} \left[ h_{\pi}^{+}(\St_{\step}, \At_{\step}) \nabla_{\paramp} \log{\pi(\At_{\step} \mid \St_{\step}, \paramp)} \right]
    \end{split}
\end{equation*}

\begin{lemma}
    \label{lem:relu-lt-abs}
    \begin{equation*}
        \text{ReLU}(a) <= |a|
    \end{equation*}
    \begin{proof}
        \begin{subequations}
            \begin{align*}
                \text{ReLU}(a) &= \max(0, a) \\
                &= \frac{1}{2}a + \frac{1}{2}|a| \\
                &= 
                \begin{cases}
                    a & a\geq0 \\
                    0 & a<0
                \end{cases} \\
                &\leq \begin{cases}
                    a & a\geq0 \\
                    -a & a<0 \quad (-a > 0)
                \end{cases} \\
                &= |a|
            \end{align*}
        \end{subequations}
    \end{proof}
    
\end{lemma}

\section{Commentary}
\label{app:commentary}

The derivation in \Cref{eq:gamma} assumes access to the policy precision parameter, $\tau = 1 / \sigma^2$, and samples of $\Ht_{\step}$. In practice, we fit $\tau$ using maximum likelihood estimation and use clipped GAEs to obtain samples of $\Ht_{\step}$. Moreover, it is only valid for continuous action spaces. We evaluate discrete action spaces below but leave theoretical grounding for future work.

Note that while we show in \Cref{eq:gamma} that approximate Bayesian inference of $\paramp$ under an assumed policy that scales actor precision, $\tau$, by clipped advantages, $h^+$, yields an equivalent likelihood objective, we do not implement a policy, $\pi(\at \mid \st, \paramp)$, that includes this scaling. We leave this exploration to future work as it requires joint inference over $\tau$ and $\paramp$ and an appropriate state conditional advantage estimator.

Finally, the conservative vector field assumption of \Cref{th:main} assumes that the actor implements a smooth function. This assumption is often broken in practice as non-smooth ReLU activation functions see use in the baselines we compare to. We leave the investigation of using smooth activation functions to future work.

\subsection{Concerning $C_{\pi}(\mathbf{s})$, $K_{\pi}$-Lipschitz Continuity, and Spectral Normalization}
\label{app:commentary-lipschitz}

In light of the dependence of the Lipschitz constant, $K_\pi$, on the policy, $\pi(\at \mid \st, \paramp)$, we offer insight into the roles played by the Lipschitz assumption and the use of critic weight spectral normalization. When we do gradient ascent according to, $$\mathbb{E}_\pi\left[ \left( q_\pi(\mathbf{S}_\mathrm{t}, \mathbf{A}_\mathrm{t}) - v_\pi(\mathbf{S}_\mathrm{t})\right)^+\nabla_{\theta} \log{\pi(\mathbf{A}_\mathrm{t} \mid \mathbf{S}_\mathrm{t}, \theta)}, \right]$$ we show that we maximize $$v^*_\pi(\mathbf{s}) \leq v_\pi(\mathbf{s}) + C_{\pi}(\mathbf{s}).$$ We want this optimization to lead to a policy $\pi$ that maximizes value, $v_\pi$, but perhaps it could lead to an undesirable policy that instead maximizes $C_{\pi}$. We show that, $$C_{\pi}(\mathbf{s}) \leq \frac{1}{2}\iint\left|v_\pi(\mathbf{s'}) - v_\pi(\mathbf{s}) \right| dP(\mathbf{s}' \mid \mathbf{S}_\mathrm{t} = \mathbf{s}, \mathbf{A}_\mathrm{t} = \mathbf{a}) d\Pi(\mathbf{a} \mid \mathbf{S}_\mathrm{t} = \mathbf{s}).$$ In theory, a policy that leads to large fluctuations in value, $v_{\pi}$, as the agent transitions from state, $\mathbf{s}$, to state, $\mathbf{s}'$, could maximize this objective.

Assuming that the value function, $v_\pi(\mathbf{s})$, is $K_\pi$-Lipschitz continuous allows us to express this bound as $$C_{\pi}(\mathbf{s}) \leq \frac{1}{2}\iint K_{\pi}\left|\left|\mathbf{s'} - \mathbf{s} \right|\right| dP(\mathbf{s}' \mid \mathbf{S}_\mathrm{t} = \mathbf{s}, \mathbf{A}_\mathrm{t} = \mathbf{a}) d\Pi(\mathbf{a} \mid \mathbf{S}_\mathrm{t} = \mathbf{s}),$$ but this does not solve the problem in itself: it could still be possible to learn a policy that merely maximizes $K_\pi$ instead of $v_\pi(\mathbf{s})$.

Hence, when we use spectral normalization of the critic weights, we regularize $K_{\pi}$ to be $1$. We find this regularization provides increased performance in most experiments run thus far. But empirically, it does not seem like the pathological behavior of maximizing $C_{\pi}(\mathbf{s})$ is happening to a significant extent even when we do not use spectral normalization. For example, we can see in \Cref{fig:rliable-mechanisms} that the performance of VSOP without spectral normalization is about equal to that of PPO on MuJoCo.

Next, we believe this analysis gives us further insight into understanding how we observe spectral normalization detrimental in highly parallel settings. In the single-threaded setting, a single agent collects data. This specific experience from a single initialization, coupled with the flexibility of Neural Networks, could result in the objective maximizing a policy that encourages spuriously high-frequency (rather than high-value) value functions when the data is sparse early in training. In this case, regularization from spectral normalization would be beneficial. Conversely, the algorithm collects data from many agents with unique initializations in the highly parallel setting. Thus, with more diverse and less sparse data, we can expect more robust value function estimates, less likely to be spuriously high-frequency between state transitions. Then, the $K_\pi=1$ assumption induced by spectral normalization may be too strong and lead to over-regularization.

\subsection{Concerning the Normal-Gamma Assumption}
\label{app:commentary-gamma}

\textbf{Is the Normal-Gamma Assumption Necessary?} The gamma-normal assumption allows us to interpret adding dropout and weight-decay regularization as sensible approximate Bayesian inference without adding complex computational overhead to the original A3C optimization algorithm. As such, this assumption primarily serves to ground Thompson sampling through approximate Bayesian inference and is not requisite for \Cref{th:main}. As with the original result of the policy gradient theorem, the results in Equations (5-6) do not make any distributional assumptions on $\pi$ and should hold for all policies with differentiable probability densities/distributions.

\begin{figure}[ht]
    \centering
    \includegraphics[width=\textwidth]{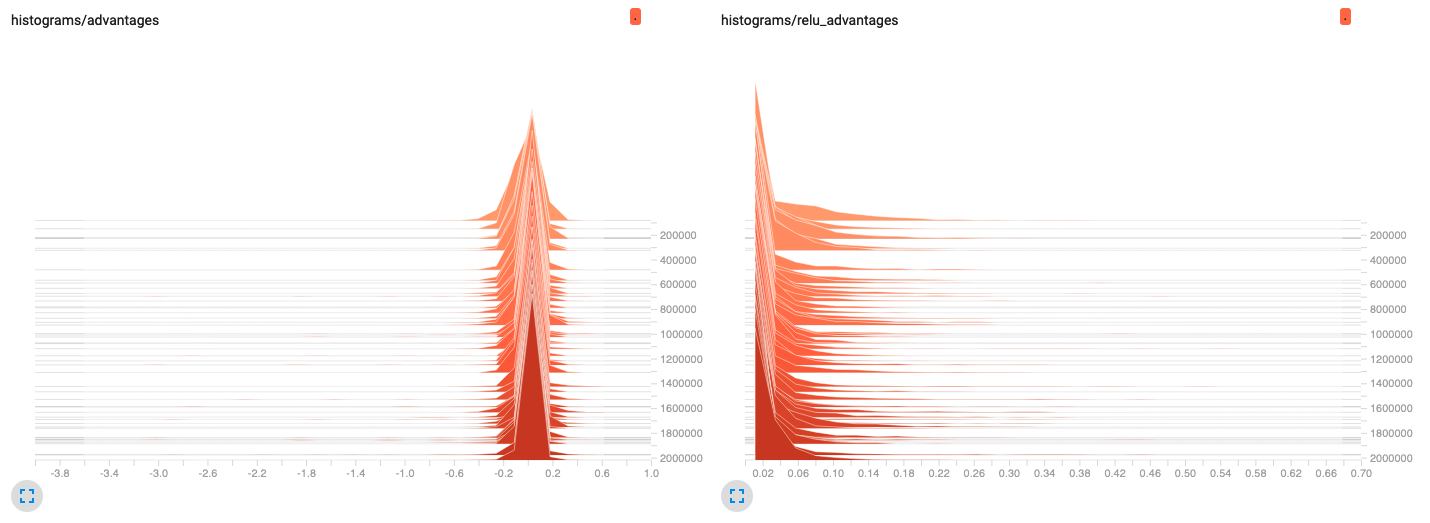}
    \caption{Comparing the histograms of estimated advantages (left) and ReLU'ed advantages (right).}
    \label{fig:advantage-histograms}
\end{figure}

\textbf{Are the Clipped Advantages Gamma Distributed?} The intuition behind assuming a gamma distribution for the clipped advantages is that advantages ideally have zero mean by construction (we subtract the state-action value by its expected state value over actions), so clipping at zero will result in a heavy-tailed distribution. Gamma distributions are sensible hypotheses for heavy-tailed distributions. In \Cref{fig:advantage-histograms} we plot the marginal histograms for the advantages (left) and the clipped advantages (right) over each training update for a training run of Humanoid-v4.

The clipped advantage histogram on the right lends evidence to the gamma assumption (at least for the marginal distribution). We may expect multi-modality at the state-action level, which integration over actions may marginalize out at the state level; however, we would still expect a heavy tail in both cases.

\section{Implementation Details}
\label{app:implementation-details}
We have attached the code that replicates the reported results in the folder ``vsop-main'' and will release a public github repo after the review process.


\subsection{Gymansium}
\label{app:implementation-details-gym}

We build off of \citet{huang2022cleanrl}'s \href{https://github.com/vwxyzjn/cleanrl}{CleanRL} package which provides reproducible, user-friendly implementations of state-of-the-art reinforcement learning algorithms using PyTorch \citep{paszke2019pytorch}, Gymnasium \citep{brockman2016openai,todorov2012mujoco}, and Weights \& Biases \citep{wandb}.
Several code-level optimizations \citep{engstrom2020implementation,andrychowicz2021matters} key to PPO reproducibility are superfluous for our method.
We omit advantage normalization, value loss clipping \citep{schulman2017proximal}, gradient clipping, and modification of the default Adam \citep{kingma2014adam} epsilon parameter as they either do not lead to an appreciable difference in performance or have a slightly negative effect. However, we find that orthogonal weight initialization, learning rate annealing, reward scaling/clipping, and observation normalization/clipping remain to have non-negligible positive effects on performance \cite{engstrom2020implementation, andrychowicz2021matters}.
In addition to adding dropout, weight decay regularization, and spectral normalization, we also look at model architecture modifications not present in the CleanRL implementation: layer width, number of hidden layers, layer activation, layer normalization \cite{ba2016layer}, and residual connections.
We find that ReLU activation functions \citep{nair2010rectified}, increasing layer width to 256, and a dropout rate of 0.01-0.04 are beneficial.
We find that network depth and residual connections are benign overall.
In contrast to recent findings in the context of offline data for off-policy reinforcement learning \citep{ball2023efficient}, layer normalization — whether applied to the actor, the critic, or both — is detrimental to performance.

\begin{table}[ht]
    \centering
    \caption{Hyper-parameters for ablation of mechinism study. VSOP, no-spectral, no-Thompson, all-actions, and no ReLU Advantage variants across Gymnasium MuJoCo environments}
    \label{table:gymnasium-ablation-hypers}
    \begin{tabular}{@{}l|ccccc@{}}
        \toprule
        & \multicolumn{5}{c}{Gymnasium MuJoCo} \\
        Parameter & VSOP & no-Spectral & all-actions & no-ReLU Adv. & no-Thompson \\ \midrule
        timesteps         & 3e6 & 3e6 & 3e6 & 3e6 & 3e6 \\
        num. envs         & 1 & 1 & 1 & 1 & 1 \\
        num. steps        & 2048 & 2048 & 2048 & 2048 & 2048 \\
        learning rate     & 2e-4 & 5.5e-4 & 2e-4 & 7.5e-4 & 2.5e-4 \\
        anneal lr         & True & True & True & True & True \\
        optim. $\epsilon$.& 1e-8 & 1e-8 & 1e-8 & 1e-8 & 1e-8 \\
        GAE $\gamma$      & 0.99 & 0.99 & 0.99 & 0.99 & 0.99 \\
        GAE $\lambda$     & 0.61 & 0.93 & 0.60 & 0.99 & 0.76 \\
        num. minibatch    & 32 & 2 & 4 & 1 & 32 \\
        update epochs     & 9 & 6 & 10 & 5 & 8 \\
        clip v-loss       & False & False & False & False & False \\
        v-loss coef.      & 0.5 & 0.5 & 0.5 & 0.5 & 0.5 \\
        max grad. norm.   & 7.1 & 8.5 & 6.4 & 8.5 & 7.2 \\
        norm. obs.        & True & True & True & True & True \\
        norm. reward      & True & True & True & True & True \\
        width             & 256 & 256 & 256 & 256 & 256 \\
        activation        & relu & relu & relu & relu & relu \\
        weight decay      & 2.4e-4 & 2.4e-4 & 2.4e-4 & 2.4e-4 & 2.4e-4 \\
        dropout           & 0.025 & 0.005 & 0.0 & 0.025 & 0.05 \\
        \bottomrule
    \end{tabular}
\end{table}

In \Cref{table:gymnasium-ablation-hypers} we present the hyperparameters used in the ablation of mechanisms study. In \Cref{table:gymnasium-hypers}, we present the hyperparameters used for the VSOP, VSPPO, RMPG, A3C, and PPO algorithms when trained on Gymnasium MuJoCo environments. The table lists hyperparameters such as the number of timesteps, thread number, and learning rate, among others. Each algorithm may have a unique set of optimal hyperparameters. Please note that some hyperparameters: 'clip $\epsilon$', 'norm. adv.', and 'clip v-loss' may not apply to all algorithms, as these are specific to certain policy optimization methods. The 'width' and 'activation' fields correspond to the architecture of the neural network used by the policy, and the 'weight decay' and 'dropout' fields pertain to the regularization techniques applied during training. In general, tuning these hyperparameters is crucial to achieving optimal performance. Note that Adam optimization \citep{kingma2014adam} is used for all algorithms except for A3C where RMSProp \citep{hinton2012rms} is used.

\begin{table}[ht]
    \centering
    \caption{Hyper-parameters for PPO, VSOP, RMPG, A3C, and VSPPO algorithms across Gymnasium MuJoCo environments}
    \label{table:gymnasium-hypers}
    \begin{tabular}{@{}l|ccccc@{}}
        \toprule
        & \multicolumn{5}{c}{Gymnasium MuJoCo} \\
        Parameter & VSOP & VSPPO & RMPG & A3C & PPO \\ \midrule
        timesteps         & 3e6 & 3e6 & 3e6 & 3e6 & 3e6 \\
        num. envs         & 1 & 1 & 1 & 1 & 1 \\
        num. steps        & 2048 & 2048 & 2048 & 5 & 2048 \\
        learning rate     & 2e-4 & 2.5e-4 & 2e-4 & 7e-4 & 3e-4 \\
        anneal lr         & True & True & True & True & True \\
        optim. $\epsilon$.& 1e-8 & 1e-8 & 1e-8 & 3e-6 & 1e-5 \\
        GAE $\gamma$      & 0.99 & 0.99 & 0.99 & 0.99 & 0.99 \\
        GAE $\lambda$     & 0.61 & 0.89 & 0.60 & 1.0 & 0.95 \\
        num. minibatch    & 32 & 64 & 4 & 1 & 32 \\
        update epochs     & 9 & 9 & 10 & 1 & 10 \\
        norm. adv.        & False & False & False & False & True \\
        clip $\epsilon$   & N/A & N/A & N/A & N/A & 0.2 \\
        clip v-loss       & False & False & False & False & True \\
        ent. coef.        & 0.0 & 0.0 & 0.0 & 0.0 & 0.0 \\
        v-loss coef.      & 0.5 & 0.5 & 0.5 & 0.5 & 0.5 \\
        max grad. norm.   & 7.1 & 2.1 & 6.4 & 0.5 & 0.5 \\
        norm. obs.        & True & True & True & True & True \\
        norm. reward      & True & True & True & True & True \\
        width             & 256 & 256 & 256 & 64 & 64 \\
        activation        & relu & relu & relu & tanh & tanh \\
        weight decay      & 2.4e-4 & 2.4e-4 & 2.4e-4 & 0.0 & 0.0 \\
        dropout           & 0.025 & 0.035 & 0.0 & 0.0 & 0.0 \\
        \bottomrule
    \end{tabular}
\end{table}

We report mean values and 95\% confidence intervals over ten random seeds.

\subsection{Gymnax}
\label{app:implementation-details-gymnax}

\begin{table}[ht]
    \centering
    \begin{tabular}{|c|c|c|c|}
        \hline
        \textbf{Hyperparameter} & \textbf{Range} & \textbf{Transformation} & \textbf{Transformed Range} \\
        \hline
        num. envs & [2, 8] & $2^{x}$ where $x$ is int & \{4, 8, 16, 32, 64, 128, 256\} \\
        \hline
        num. steps & [2, 8] & $2^{x}$ where $x$ is int & \{4, 8, 16, 32, 64, 128, 256\} \\
        \hline
        $\lambda$ & [0.0, 1.0] & round to multiple of 0.002 & \{0.0, 0.002, \ldots, 1.0\} \\
        \hline
        learning rate & [1e-4, 1e-3] & round to multiple of 0.00005 & \{1e-4, 1.5e-5, \ldots, 1e-3\} \\
        \hline
        max grad. norm. & [0.2, 5.0] & round to multiple of 0.1 & \{0.2, 0.3, \ldots, 5.0\} \\
        \hline
        num. minibatch & [0, 6] & $2^{x}$ where $x$ is int & \{1, 2, 4, 8, 16, 32, 64\} \\
        \hline
        update epochs & [1, 10] & round to int & \{1, 2, 3, ..., 10\} \\
        \hline
        width & [6, 10] & $2^{x}$ where $x$ is int & \{64, 128, 256, 512, 1024\} \\
        \hline
    \end{tabular}    \caption{Hyperparameter search space with transformations}
    \label{table:search_space_transformed}
\end{table}

We optimize the hyper-parameters for each algorithm for each set of environments using a Bayesian optimization search strategy \citep{snoek2012practical}. Each algorithm has a budget of 100 search steps. We use NVIDIA A100 GPUs. The hyperparameters we search over include learning rate, number of steps, number of environments,  GAE $\lambda$, update epochs, number of minibatches, and the maximum gradient norm. We also search over the hidden layer width for Brax-MuJoCo and MinAtar environments. Each hyperparameter has a specific search space and transformation applied during the search. We summarize the search sapce in \Cref{table:search_space_transformed}.

For the MinAtar environments, the hyper-parameters search spaces are: the number of steps in $[2, 8]$ (transformed to $2^{x}$ where $x$ is the integer part of the sample), GAE $\lambda$ in $[0.0, 1.0]$ (rounded to the nearest multiple of $0.002$), learning rate in $[1e-4, 1e-3]$ (rounded to the nearest multiple of $0.00005$), update epochs in $[1, 10]$ (rounded to the nearest integer), maximum gradient norm in $[0.0, 5.0]$ (rounded to the nearest multiple of $0.1$), number of minibatches in $[0, 6]$ (transformed to $2^{x}$), update epochs in $[1, 10]$ (rounded to the nearest integer), and number of minibatches in $[0, 7]$ (transformed to $2^{x}$), and hidden layer width in $[6, 10]$ (transformed to $2^{x}$). We set the $\gamma$ and number of environments to fixed values at $0.99$ and $64$, respectively.

For MuJoCo-Brax, we do not search over the number of environments or steps. Instead we set them to fixed values at $0.99$, $2048$, and either $10$ or $5$, respectively. The search space for the remaining hyper-parameters the same ranges as for the MinAtar environments. Further, we only optimize over the Humanoid, Hopper, and Reacher environments for 20 million steps. We test for each environment for 50 million steps.

Finally, for Classic Control environments, we employ the same hyperparameter search as for MinAtar, except that we search over the number of environments in $[2, 8]$ (transformed to $2^{x}$ where $x$ is the integer part of the sample) and we do not search over the hidden layer width, instead setting it to a fixed value of $64$.

This strategy allows us to thoroughly explore the hyperparameter space and find values that generalize well across a variety of different tasks. Further it allows us to fairly compare each algorithm. \Cref{table:minatar-hypers,table:brax-mujoco-hypers,table:classic-control-hypers} report the final hyper-parameter values for PPO, VSOP, and A3C.

\begin{table}[ht]
    \centering
    \caption{PPO, VSOP, A3C, and DPO Hyper-parameters for  MinAtar environments.}
    \label{table:minatar-hypers}
    \begin{tabular}{@{}l|cccc@{}}
        \toprule
        Parameter       & PPO & VSOP & A3C & DPO \\ \midrule
        learning rate   & 9e-4 & 7.5e-4 & 7e-4 & 1e-3 \\
        num. envs       & 128 & 128 & 128 & 128 \\
        num. steps      & 64 & 32 & 4 & 16 \\
        GAE $\gamma$    & 0.99 & 0.99 & 0.99 & 0.99  \\
        GAE $\lambda$   & 0.70 & 0.82 & 0.87 & 0.70 \\
        num. minibatch  & 8 & 16 & 2 & 8 \\
        update epochs   & 10 & 9 & 1 & 6 \\
        max grad. norm. & 1.9 & 2.8 & 1.3 & 0.4 \\
        width           & 512 & 512 & 512 & 256 \\
        activation      & relu & relu & relu & relu \\
        clip $\epsilon$ & 0.2 & N/A & N/A & 0.2 \\
        ent. coef.      & 0.01 & 0.01 & 0.01 & 0.01 \\
        \bottomrule
    \end{tabular}
\end{table}

\begin{table}[ht]
    \centering
    \caption{Hyper-parameters for PPO, VSOP, A3C, and DPO algorithms across Brax-MuJoCo environments}
    \label{table:brax-mujoco-hypers}
    \begin{tabular}{@{}l|cccc@{}}
        \toprule
        Parameter       & PPO & VSOP & A3C & DPO \\ \midrule
        learning rate   & 4.5e-4 & 1e-4 & 7e-4 & 2e-4 \\
        num. envs       & 2048 & 2048 & 2048 & 2048 \\
        num. steps      & 10 & 10 & 5 & 10 \\
        GAE $\gamma$    & 0.99 & 0.99 & 0.99 & 0.99 \\
        GAE $\lambda$   & 0.714 & 1.0 & 0.97 & 0.942 \\
        num. minibatch  & 32 & 64 & 2 & 32  \\
        update epochs   & 3 & 2 & 1 & 6 \\
        max grad. norm. & 3.3 & 3.7 & 1.0 & 0.4 \\
        width           & 512 & 512 & 128 & 512 \\
        activation      & relu & relu & relu & relu\\
        clip $\epsilon$ & 0.2 & N/A & N/A & 0.2 \\
        ent. coef.      & 0.0 & 0.0 & 0.0 & 0.0 \\
        \bottomrule
    \end{tabular}
\end{table}

\begin{table}[ht]
    \centering
    \caption{Hyper-parameters for PPO, VSOP, A3C, and DPO algorithms across Classic Control environments}
    \label{table:classic-control-hypers}
    \begin{tabular}{@{}l|cccc@{}}
        \toprule
        Parameter & PPO & VSOP & A3C & DPO \\ \midrule
        learning rate   & 1e-3 & 8.5e-4 & 5.5e-4 & 1e-3 \\
        num. envs       & 8 & 16 & 8 & 4 \\
        num. steps      & 8 & 64 & 4 & 4 \\
        GAE $\gamma$    & 0.99 & 0.99 & 0.99 & 0.99 \\
        GAE $\lambda$   & 0.54 & 0.58 & 0.13 & 1.0 \\
        num. minibatch  & 8 & 16 & 8 & 1 \\
        update epochs   & 3 & 8 & 1 & 10 \\
        max grad. norm. & 3.4 & 1.9 & 3.8 & 5.0 \\
        width           & 64 & 64 & 64 & 64 \\
        activation      & tanh & tanh & tanh & tanh \\
        clip $\epsilon$ & 0.2 & N/A & N/A & 0.2 \\
        ent. coef.      & 0.01 & 0.01 & 0.01 & 0.01 \\
        \bottomrule
    \end{tabular}
\end{table}

All reported results for MinAtar, Classic Control, and MuJoCo-Brax respectively are given by mean values and 68\% confidence intervals over 20 random seeds. During tuning we use 2 random seeds and for testing we use a different set of 20 random seeds, as per the guidance of \citet{eimer2023hyperparameters}.

\subsection{ProcGen}
\label{app:implementation-details-procgen}

ProcGen \citep{cobbe2020leveraging} is a set of 16 environments where game levels are procedurally generated, creating a virually unlimited set of unique levels. 
We follow the ``easy'' generalization protocol where, for a given environment, models are trained on 200 levels for 25 million time steps and evaluated on the full distribution of environments.
We use the same architecture as PPO in the CleanRL library \citep{huang2022cleanrl}, and do a Bayesian optimization hyper-parameter search \citep{snoek2012practical} using the bossfight environment.
We search over the learning rate, GAE $\lambda$, number of minibatches per epoch, number of epochs per rollout, the dropout rate, and the entropy regularization coefficient. We report the final VSOP hyperparamters in \Cref{table:procgen-hypers} and include the relevant PPO hyperparameters for comparison. Note also that, VSOP does not make use of advantage normalization or value loss clipping.

\begin{table}[]
    \centering
    \begin{tabular}{lcccccc}
        \toprule
         Method & lr & GAE $\lambda$ & num. minibatch & update epochs & dropout & ent. coef. \\
         \midrule
         VSOP & 4.5e-4 & 0.88 & 8 & 3 & 0.075 & 1e-5 \\
         PPO & 5.0e-4 & 0.95 & 8 & 3 & 0.000 & 1e-2 \\
        \bottomrule
    \end{tabular}
    \caption{Final ProcGen hyperparameters for VSOP}
    \label{table:procgen-hypers}
\end{table}

\section{Additional Results}
\label{app:additional-results}

\subsection{Comparison to Baselines}
\label{app:additional-results-baselines}

\begin{figure}[ht]
    \centering
    \includegraphics[width=\textwidth]{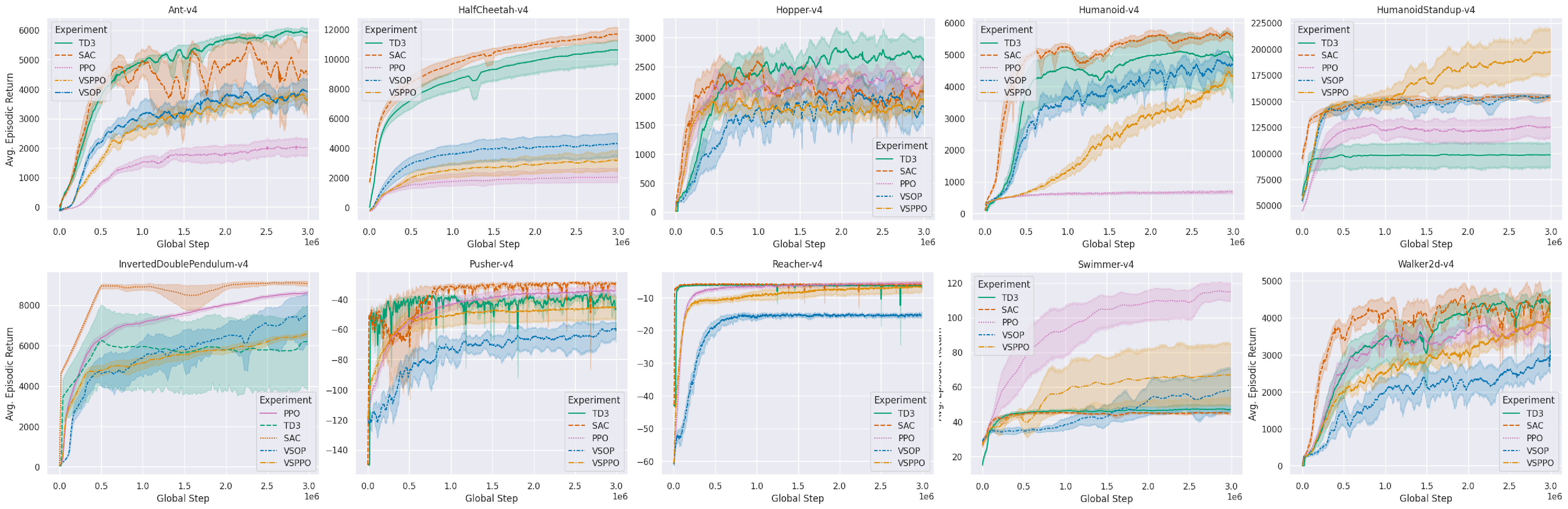}
    \caption{
        Gymnasium-MuJoCo. Comparing VSOP to baseline algorithms.
    }
    \label{fig:mujoco-baselines}
\end{figure}

\Cref{fig:mujoco-baselines} compares VSOP training curves to baseline algorithms.

\subsection{Spectral Normalization and Thompson Sampling May Improve PPO}
\label{app:ppo}
\label{sec:exp-gymnasium-ppo}
\begin{figure}[ht]
    \centering
    \includegraphics[width=\textwidth]{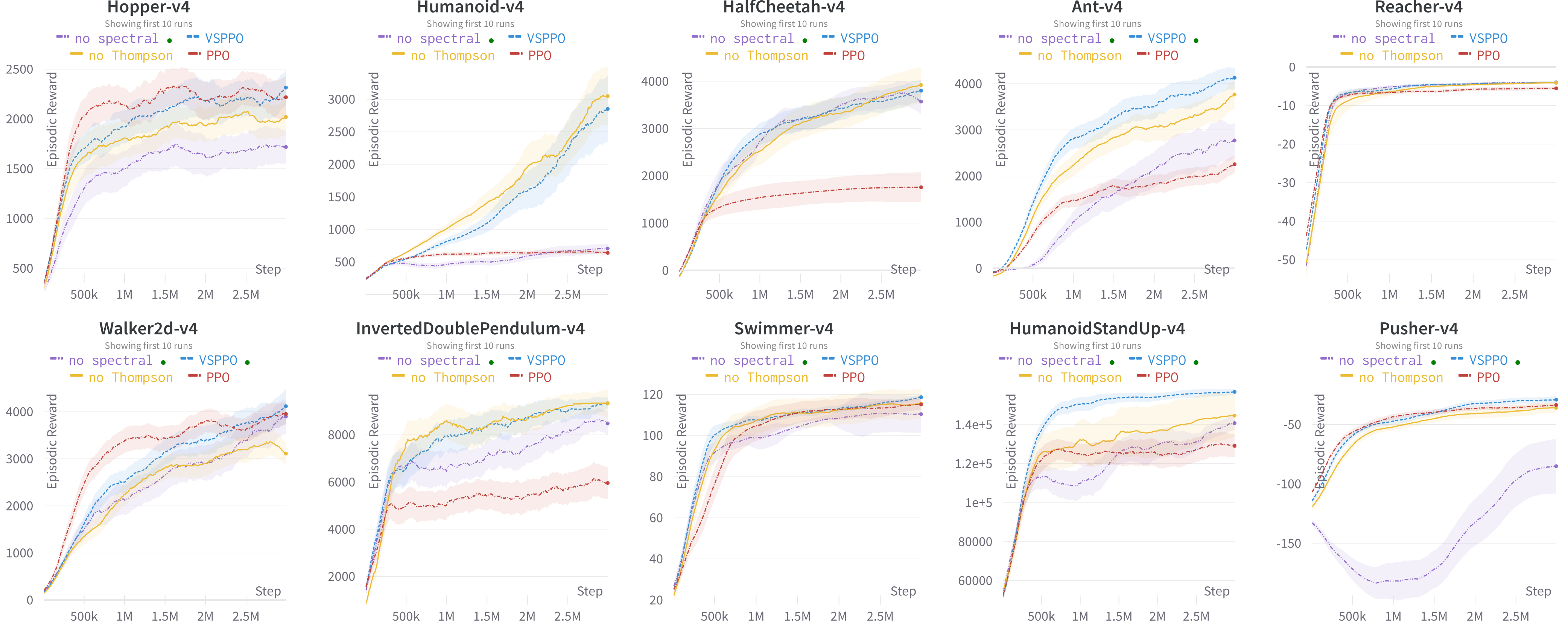}
    \caption{MuJoCo continuous control benchmark examining the effect of Thompson sampling and spectral normalization on PPO.}
    \label{fig:mujoco-vsppo}
\end{figure}

Interestingly, we see this same trend when applying spectral normalization and dropout to PPO. 
In \Cref{fig:mujoco-vsppo} we compare how Thompson sampling and spectral norm effect PPO.

\subsection{Gymnax Environments}
\label{sec:exp-gymnax}
PureJaxRL \citep{lu2022discovered} uses Gymnax \citep{gymnax2022github} and Jax \citep{jax2018github} to enable vectorization, which facilitates principled hyper-parameter tuning.
Using it, we explore several environments and compare VSOP, PPO, A3C, and DPO. We use Bayesian hyper-parameter optimization \citep{snoek2012practical} and give each algorithm a search budget of 100 steps. We search over hyper-parameters such as the learning rate, number of update epochs, number of mini-batches in an update epoch, the GAE $\lambda$ parameter, the max gradient norm, and the width of the network. We give full implementation details in \Cref{app:implementation-details-gymnax}. \Cref{tab:gymnax-ranking} shows the overall ranking of each method. VSOP is competitive with DPO and improves over PPO and A3C.

\begin{table}[ht]
    \caption{Rank scores (lower is better) for VSOP, DPO, PPO, and A3C on Brax-MuJoCo, MinAtar, and Classic Control. Methods are ranked from 1 to 4 based on statistically significant differences (paired t-test with p-value 0.1) between mean last episode returns. Ties are given the same rank, and the proceeding score will be the last rank plus the number of additional methods.}
    \label{tab:gymnax-ranking}
    \centering
    \begin{tabular}{l|ccc|c}
        \toprule
        \textbf{Method} & \textbf{Brax-MuJoCo} & \textbf{MinAtar} & \textbf{Classic Control} & \textbf{Avg. Rank} \\
        \midrule
        DPO         & \textbf{1.33} & \textbf{1.75} & 1.25 & 1.44 \\
        VSOP (Ours) & 1.78 & 2.50 & \textbf{1.00} & 1.76 \\
        PPO         & 2.00 & 2.25 & 1.25 & 1.83 \\
        A3C         & 4.00 & 2.25 & 1.25 & 2.50 \\
        \bottomrule
    \end{tabular}
\end{table}

\begin{figure}[ht]
    \centering
    \begin{subfigure}[b]{0.24\textwidth}
        \centering
        \includegraphics[width=\textwidth]{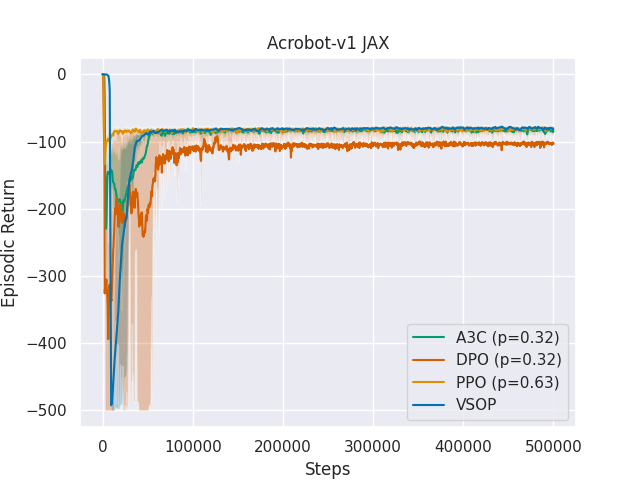}
        \caption{Acrobot}
        \label{fig:classic-control-acrobot}
    \end{subfigure}
    \hfill
    \begin{subfigure}[b]{0.24\textwidth}
        \centering
        \includegraphics[width=\textwidth]{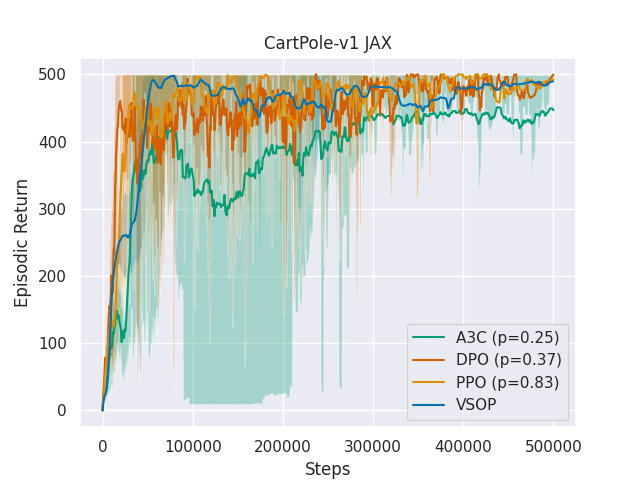}
        \caption{CartPole}
        \label{fig:classic-control-cartpole}
    \end{subfigure}
    \hfill
    \begin{subfigure}[b]{0.24\textwidth}
        \centering
        \includegraphics[width=\textwidth]{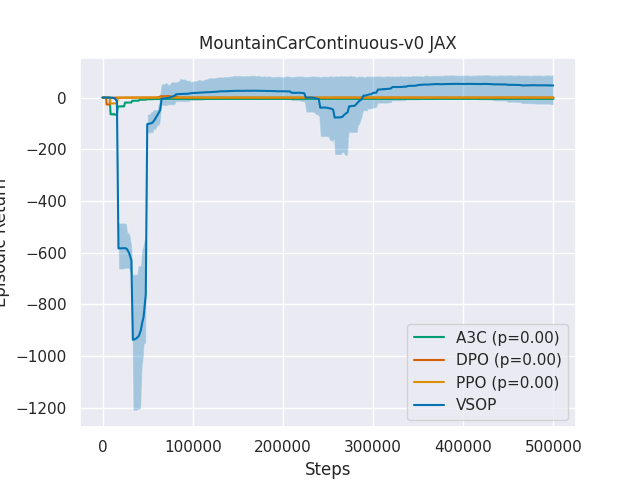}
        \caption{MountainCar Cont.}
        \label{fig:classic-control-mountaincar}
    \end{subfigure}
    \hfill
    \begin{subfigure}[b]{0.24\textwidth}
        \centering
        \includegraphics[width=\textwidth]{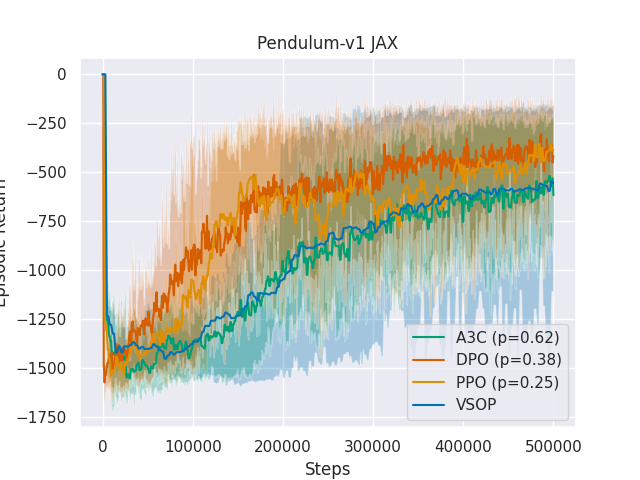}
        \caption{Pendulum}
        \label{fig:classic-control-pendulum}
    \end{subfigure}
    \caption{
        Classic Control Environments \citep{gymnax2022github}. Mean episodic return and 68\% CI over 20 random seeds are shown for VSOP (Blue), PPO (Orange), A3C (Green), and DPO (Red). Each method is hyper-parameter tuned using Bayesian Optimization with 100 search steps. Paired t-test p-values for last episode with respect to VSOP shown in brackets. Significant improvement is seen for VSOP compared to all other methods on MountainCar Continuous.
    }
    \label{fig:classic-control}
\end{figure}

\Cref{fig:classic-control} summarize the results for \textbf{Classic Control}.
Performance of each method is in general statistically equal, but VSOP shows significant gain on MountainCar Continuous.

\begin{figure}[ht]
    \centering
    \begin{subfigure}[b]{0.24\textwidth}
        \centering
        \includegraphics[width=\textwidth]{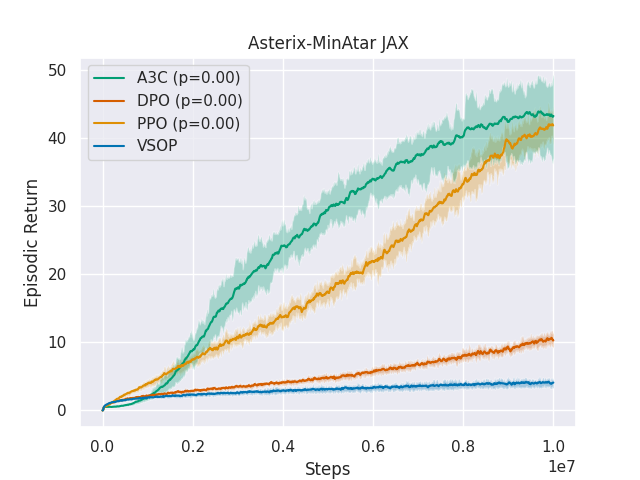}
        \caption{Asterix}
        \label{fig:minatar-asterix}
    \end{subfigure}
    \hfill
    \begin{subfigure}[b]{0.24\textwidth}
        \centering
        \includegraphics[width=\textwidth]{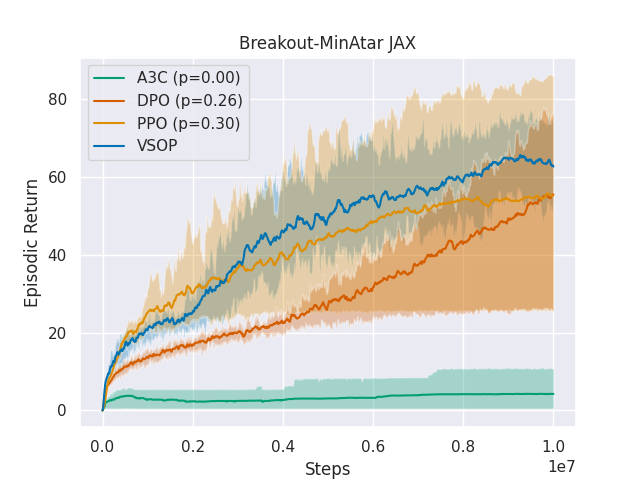}
        \caption{Breakout}
        \label{fig:minatar-breakout}
    \end{subfigure}
    \hfill
    \begin{subfigure}[b]{0.24\textwidth}
        \centering
        \includegraphics[width=\textwidth]{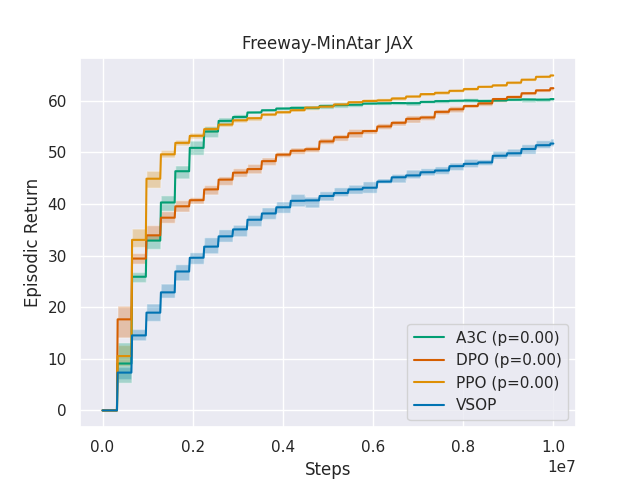}
        \caption{Freeway}
        \label{fig:minatar-freeway}
    \end{subfigure}
    \hfill
    \begin{subfigure}[b]{0.24\textwidth}
        \centering
        \includegraphics[width=\textwidth]{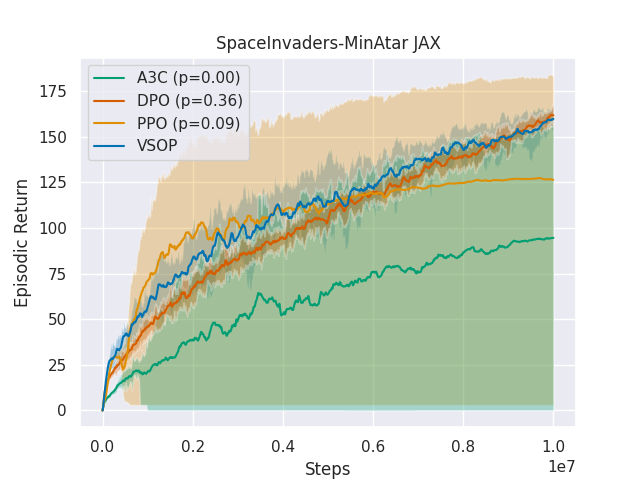}
        \caption{SpaceInvaders}
        \label{fig:minatar-spaceinvaders}
    \end{subfigure}
    \caption{
        MinAtar Environments \citep{young19minatar}. Mean episodic return and 68\% CI over 20 random seeds are shown for VSOP (Blue), PPO (Orange), A3C (Green), and DPO (Red). Methods are hyper-parameter tuned using Bayesian Optimization with 100 search steps. p-values for last episode with respect to VSOP shown in brackets. VSOP performs well on Breakout and SpaceInvaders.
    }
    \label{fig:minatar}
\end{figure}

\Cref{fig:minatar} summarize the results for \textbf{MinAtar} \citep{bellemare2013arcade,young19minatar}. VSOP shows significant improvement over PPO and A3C in Space Invaders. We see marginal improvement over PPO and DPO in Breakout, with significant improvement over A3C. VSOP trails the baselines significantly in Asterix and Freeway.

\begin{figure}[ht]
    \centering
    \begin{subfigure}[b]{0.3\textwidth}
        \centering
        \includegraphics[width=\textwidth]{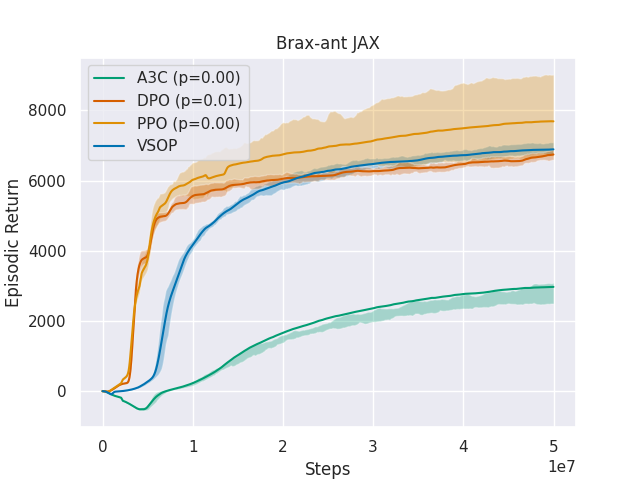}
        \caption{Brax-ant}
        \label{fig:Brax-ant}
    \end{subfigure}
    \hfill
    \begin{subfigure}[b]{0.3\textwidth}
        \centering
        \includegraphics[width=\textwidth]{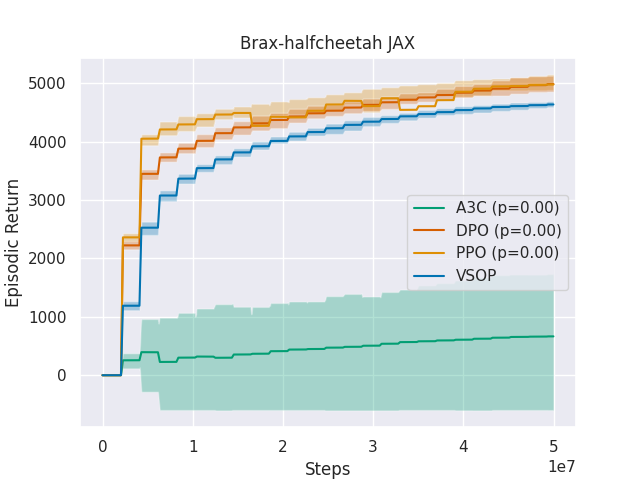}
        \caption{Brax-halfcheetah}
        \label{fig:Brax-halfcheetah}
    \end{subfigure}
    \hfill
    \begin{subfigure}[b]{0.3\textwidth}
        \centering
        \includegraphics[width=\textwidth]{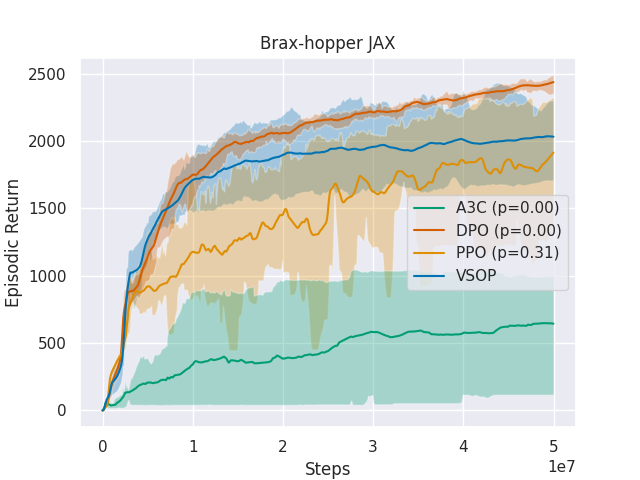}
        \caption{Brax-hopper}
        \label{fig:Brax-hopper}
    \end{subfigure}
    
    \begin{subfigure}[b]{0.3\textwidth}
        \centering
        \includegraphics[width=\textwidth]{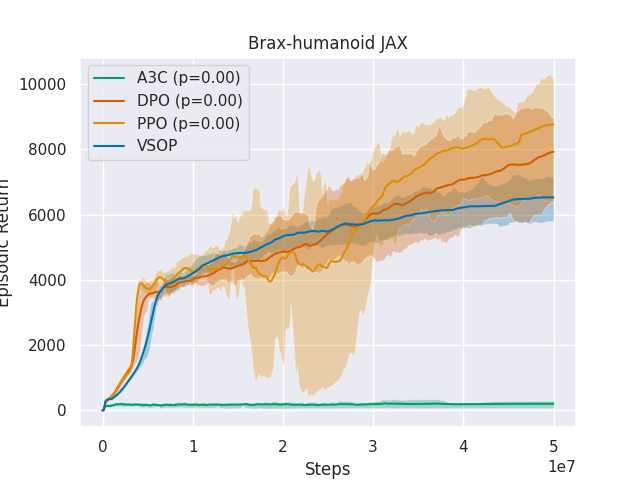}
        \caption{Brax-humanoid}
        \label{fig:Brax-humanoid}
    \end{subfigure}
    \hfill
    \begin{subfigure}[b]{0.3\textwidth}
        \centering
        \includegraphics[width=\textwidth]{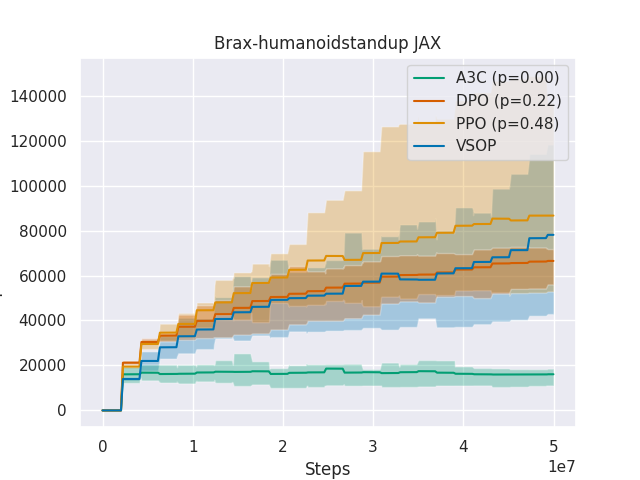}
        \caption{Brax-humanoidstandup}
        \label{fig:Brax-humanoidstandup}
    \end{subfigure}
    \hfill
    \begin{subfigure}[b]{0.3\textwidth}
        \centering
        \includegraphics[width=\textwidth]{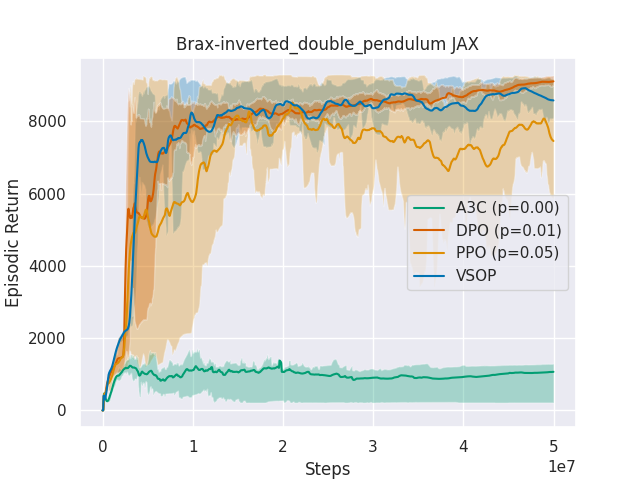}
        \caption{Brax-doublependulum}
        \label{fig:Brax-inverted_double_pendulum}
    \end{subfigure}
    
    \begin{subfigure}[b]{0.3\textwidth}
        \centering
        \includegraphics[width=\textwidth]{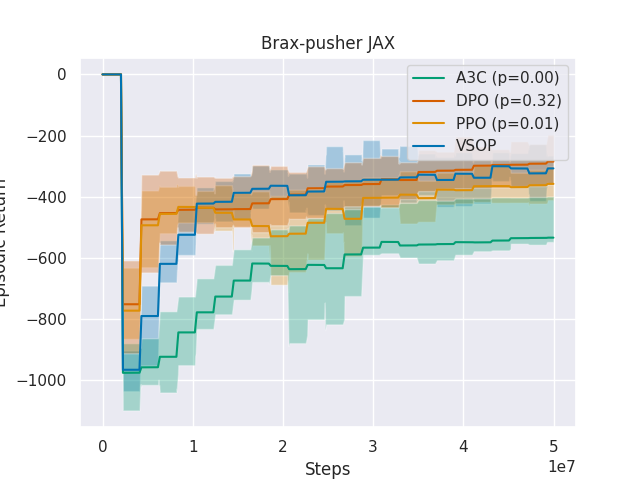}
        \caption{Brax-pusher}
        \label{fig:Brax-pusher}
    \end{subfigure}
    \hfill
    \begin{subfigure}[b]{0.3\textwidth}
        \centering
        \includegraphics[width=\textwidth]{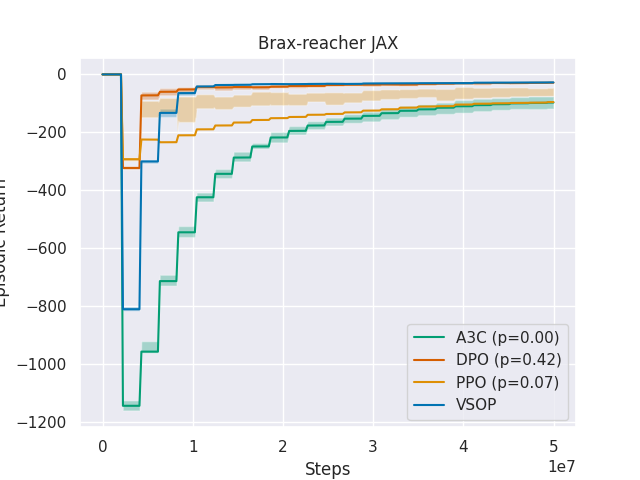}
        \caption{Brax-reacher}
        \label{fig:Brax-reacher}
    \end{subfigure}
    \hfill
    \begin{subfigure}[b]{0.3\textwidth}
        \centering
        \includegraphics[width=\textwidth]{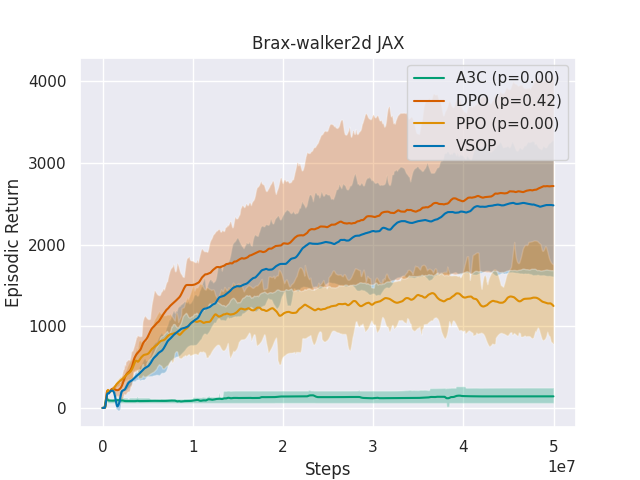}
        \caption{Brax-walker2d}
        \label{fig:Brax-walker2d}
    \end{subfigure}
    \caption{
        Brax-MuJoCo Environments \citep{brax2021github,todorov2012mujoco}. Mean episodic return and 68\% CI over 20 random seeds are shown for VSOP (Blue), PPO (Orange), A3C (Green), and DPO (red). Each method is hyper-parameter tuned using Bayesian Optimization \citep{snoek2012practical} with a budget of 100 search steps. Paired t-test p-values for last episode with respect to VSOP shown in brackets. VSOP generally out performs PPO and A3C and is competitive with DPO.
    }
    \label{fig:brax}
\end{figure}

\Cref{fig:brax} summarize the results for \textbf{Brax MuJoCo} \citep{todorov2012mujoco,brax2021github}. We perform paired t-tests for the last episode between each method and VSOP. We threshold at a p-value of 0.1 to indicate significance. VSOP significantly outperforms A3C in all environments. VSOP significantly outperforms PPO in four of nine environments (InvertedDoublePendulum, Pusher, Reacher, and Walker2d), is statistically equivalent in two environments (Hopper and HumanoidStandUp), and is significantly less effective in three environments (Ant, HalfCheetah, and Humanoid). VSOP outperforms DPO on Ant, is statistically equivalent in four environments (HumanoidStandUp, Pusher, Reacher, and Walker2d), but is significantly less effective in four environments (HalfCheetah, Hopper, Humanoid, and InvertedDoublePendulum). Overall, VSOP outperforms A3C and PPO and is competitive with DPO.

\end{document}